\documentclass[runningheads]{llncs}
\usepackage{url}
\usepackage[bottom]{footmisc}
\usepackage[caption=false,font=footnotesize]{subfig}
\usepackage{epsfig}
\usepackage{amsmath}
\usepackage{amssymb}
\usepackage{amsfonts}
\usepackage{breakcites}
\usepackage{array}

\newcolumntype{P}[1]{>{\centering\arraybackslash}p{#1}}
\graphicspath{{./images/}}

\usepackage{graphicx}
\usepackage[graphicx]{realboxes}

\usepackage{tikz}
\usepackage{comment}
\usepackage{amsmath,amssymb} 
\usepackage{color, colortbl}
\usepackage{multirow}
\usepackage{hyperref}
\usepackage{xcolor}
\hypersetup{
    colorlinks,
    linkcolor={black!50!black},
    citecolor={black!50!black},
    urlcolor={black!80!black},
}
\usepackage{listings}
\usepackage{booktabs}
\usepackage[normalem]{ulem}
\useunder{\uline}{\ul}{}
\usepackage{algpseudocode}
\usepackage{algorithm}
\algnewcommand\algorithmicforeach{\textbf{for each}}
\algnewcommand{\algorithmicand}{\textbf{ and }}
\algnewcommand{\AND}{\algorithmicand}
\algdef{S}[FOR]{ForEach}[1]{\algorithmicforeach\ #1\ \algorithmicdo}
\algnewcommand{\algorithmicor}{\textbf{ or }}
\algnewcommand{\OR}{\algorithmicor}
\algnewcommand\algorithmicnot{\textbf{not}}
\algdef{SE}[IF]{IfNot}{EndIf}[1]{\algorithmicif\ \algorithmicnot\ #1\ \algorithmicthen}{\algorithmicend\ \algorithmicif}%

\definecolor{dkgreen}{rgb}{0,0.6,0}
\definecolor{gray}{rgb}{0.5,0.5,0.5}
\definecolor{mauve}{rgb}{0.58,0,0.82}

\definecolor{top_3}{HTML}{E8E9EC}
\definecolor{top_2}{HTML}{d9d9d9}
\definecolor{top_1}{HTML}{BFBFBF}

\newcommand{\currier}[1]{{\fontfamily{qcr}\selectfont #1}}
\newcommand{\comm}[1]{}

\begin{document}
\title{A Comprehensive Gold Standard and Benchmark for Comics Text Detection and Recognition}
\titlerunning{COMICS TEXT+}
%
\author{Gürkan Soykan\inst{1} \and
    Deniz Yuret\inst{1} \and
    Tevfik Metin Sezgin\inst{1}}
\authorrunning{G. Soykan et al.}
%
\institute{KUIS AI Center, Koç University
    \url{https://ai.ku.edu.tr/} \\
    \email{\{gsoykan20,dyuret,mtsezgin\}@ku.edu.tr}}
\maketitle              

\begin{abstract}
    This study focuses on improving the optical character recognition (OCR) data for panels in the COMICS dataset \cite{iyyer2017amazing}, the largest dataset containing text and images from comic books. To do this, we developed a pipeline for OCR processing and labeling of comic books and created the first text detection and recognition datasets for western comics, called \textit{"COMICS Text+: Detection"} and \textit{"COMICS Text+: Recognition"}. We evaluated the performance of state-of-the-art text detection and recognition models on these datasets and found significant improvement in word accuracy and normalized edit distance compared to the text in COMICS. We also created a new dataset called \textit{"COMICS Text+"}, which contains the extracted text from the textboxes in the COMICS dataset. Using the improved text data of COMICS Text+ in the comics processing model from \cite{iyyer2017amazing} resulted in state-of-the-art performance on cloze-style tasks without changing the model architecture. The \textit{COMICS Text+} dataset can be a valuable resource for researchers working on tasks including text detection, recognition, and high-level processing of comics, such as narrative understanding, character relations, and story generation.
    All the data and inference instructions can be accessed in
    \href{https://github.com/gsoykan/comics_text_plus}{https://github.com/gsoykan/comics\textunderscore text\textunderscore plus}.

\end{abstract}

\section{Introduction}

Comics are a multimodal structure and medium that use writing and images in a spectrum, from purely images to purely writing, to convey a story or an idea. Comics are at the intersection of education \cite{education_herbst2011using}, sociocultural studies \cite{sociological_approach_brienza2010producing}, linguistics and cognitive sciences \cite{computational_approaches_laubrock2020computational,comicsoverview_augereau2017overview}. Understanding comics can help advance progress in these fields. Additionally, methods developed for processing and understanding comics can be applied to other forms of "Visual Language," including but not limited to cave paintings, mangas, graphic novels, and comics \cite{visual_language_of_comics_cohn2013visual}.

The unique structure of comics can be analyzed at the low-level and high-level. At the low-level, natural language processing techniques can extract text from dialogues, narratives, and onomatopoeias. Computer vision methods can detect and segment motion lines, characters, and the location of objects and components \cite{comic_mtl_nguyen2019comic,balloon_assoc_rigaud2015speech,ocrpipeline_hartel2021ocr,dcm772dataset_digital_comic_indexing}. At the high-level, the relationships between characters, storytelling, narrative understanding, inter-panel and intra-panel events must be studied. While the high-level features of natural images and scenes are widely studied \cite{minaee2021image}, the same cannot be said for comics. This is due to the great variety of comics across time, location, and different styles and the lack of annotated data (see Table \ref{table:datasets_overview} for an overview of datasets). The limited availability of datasets in this domain can also be attributed to copyright and access rights challenges.

The amount of annotated data creates a bottleneck for studies on comics using deep learning methods. Although there have been works on panel and speech bubble detection and segmentation, such as in \cite{comic_mtl_nguyen2019comic,balloon_assoc_rigaud2015speech,ocrpipeline_hartel2021ocr,dcm772dataset_digital_comic_indexing}, and onomatopoeia detection and recognition, such as in \cite{onomatopoeia_baek2022coo}, the only extensive dataset with OCR-extracted comics texts is called COMICS \cite{iyyer2017amazing}. However, when the text quality of the COMICS dataset is measured over a sample, it can be seen as unreliable (see Subsection \ref{subsection:comics_flaws} and Table \ref{table:final_results_comics_vs_ours}). There is a limited amount of study on the OCR of dialogue and narrative texts in comics to produce high-quality text data. In \cite{ocrpipeline_hartel2021ocr}, such a pipeline was presented, but the data and models could not be shared due to copyright issues.

This study presents a pipeline for OCR of comics, allowing us to extract text from speech bubbles in panels. This pipeline can also be used for general OCR purposes. We use the COMICS dataset \cite{iyyer2017amazing} to curate a reliable and comprehensive dataset of American golden age comics. These comics are uniform, allowing our work to be directly applied to previous studies in the literature. We inspect the reported and unreported problems in the COMICS dataset and develop separate solutions for text detection and recognition using the MMOCR toolbox \cite{mmocr2021}. To improve the performance of text detection and recognition, we follow a loop of annotating panels and speech bubbles, training models and evaluating them using ground-truth data. Once we have achieved stable performance with the trained models, we select the best end-to-end text detection and recognition model pair to process the COMICS dataset and extract OCR texts. Finally, we apply post-processing techniques to correct systematic errors in the OCR data and create the \textit{COMICS TEXT+} dataset. Our contributions can be summarized as follows:

\begin{itemize}
    \item We release a substantially improved version of the COMICS dataset OCR results, called \textit{COMICS Text+}, which significantly outperforms its predecessor in terms of text quality, with a word accuracy improvement from 0.13 to 0.40. It is the best quality and most comprehensive publicly available dataset, with over two million transcriptions of textboxes.
    \item We release text detection and text recognition datasets created from Golden Age comics called respectively, \textit{COMICS Text+: Detection} and \textit{COMICS Text+: Recognition} along with ground truth data to validate end-to-end OCR pipelines. \textit{COMICS Text+: Detection} contains more than 20,000 annotations from 1112 images. Whereas, \textit{COMICS Text+: Recognition} contains more than 17,000 annotations from 1006 images.
    \item We train and evaluate the performance of 14 text detection and 10 text recognition models on COMICS Text+ datasets and analyze the effects of training size on performance. Our results show that text detection models are the bottleneck for OCR studies of comics, (see Figure \ref{fig:best_det_data_size}) but their performance improves significantly after 200 textbox annotations.
    \item We present fine-tuned text detection and recognition models, based on FCENet \cite{FCENET_zhu2021fourier} and MASTER \cite{Lu2021MASTER}, respectively, that have been trained on the \textit{COMICS Text+: Detection \& Recognition} datasets. There are no other publicly available models and datasets of this specific type, so we hope this work will serve as a baseline for future studies and lower the barrier to entry for those interested in this domain.
    \item We share an annotation tool forked from LabelMe \cite{wada2018labelme}. Since LabelMe does not allow the use of models to get predictions, we improved it by enabling it to use text detection and text recognition models to aid and speed up the annotation process. Our tool converts annotations into text detection, and recognition datasets used to create \textit{COMICS Text+: Detection \& Recognition} datasets. By providing this tool, we hope to make it easier for other researchers to create high-quality datasets for training and evaluating OCR models and can help to advance the state of the art in this field.
    \item The comics processing backbone for cloze-style tasks proposed in the COMICS dataset is reproduced, and the reproduced model is trained using both COMICS and \textit{COMICS TEXT+} datasets. The test results of the two models are compared, and it is shown that the model trained with \textit{COMICS TEXT+} outperforms the other model and achieves state-of-the-art results on most cloze-style tasks.
\end{itemize}

\section{Related Work}
\label{section:rel_works}

\subsection{Text Detection}

Deep learning techniques have recently been widely used for scene text detection. These techniques can be broadly classified into three types based on the granularity of the prediction results: regression-based, part-based, and segmentation-based approaches \cite{DBNETPP_liao2022real}.

Regression-based approaches directly predict each instance's bounding box coordinates and use simple post-processing techniques (e.g., non-maximum suppression) to generate final results. However, these approaches may not perform well when dealing with irregularly shaped objects, such as curved text \cite{liao2017textboxes,liao2018textboxes++,FCENET_zhu2021fourier,xie2019derpn}.

Part-based methods detect smaller parts of a text (e.g., words) and link them together to form larger units (e.g., sentences). These methods are effective for detecting long texts, but their linking algorithms can be complex and require hand-tuned parameters, making them difficult to optimize \cite{seglink_shi2017detecting,tang2019seglink++}.

Segmentation-based methods make predictions on the pixel level and use component-grouping paradigms as post-processing algorithms to combine the predictions and generate mask outputs or bounding boxes \cite{DBNETPP_liao2022real, PSENET_wang2019shape,masktextspotter_lyu2018mask}. These methods can be more accurate than regression-based and part-based approaches, but they may be computationally intensive.

\subsection{Text Recognition}

A text recognition system takes an image of a text instance and converts it into a sequence of characters. Scene text recognition systems are a special type of text recognizer designed to handle text instances that come from natural scenes. This type of text recognition differs from document-level OCR systems because images from natural scenes often have complex backgrounds, varying fonts, changing colors, and poor imaging conditions \cite{textrecognitionsurvey_chen2021text}. These challenges are similar to those encountered in comics, which is why we focus on scene text recognition systems in our study.

Like other branches of computer vision, text recognition systems have moved from using hand-crafted features to deep learning-based models. In ABINet \cite{ABINet-fang2021read}, a combination of vision models and explicitly defined language models is used for scene text recognition. The language model acts as a simulator for the cloze test, and an iterative correction strategy is applied to the visual model to improve the results. Another scene text recognizer we use in our study is MASTER \cite{Lu2021MASTER}, which can learn feature-feature and target-target relationships within its encoder-decoder structure, resulting in better intermediate representations.

\subsection{OCR for Comics}

In a preliminary study, \cite{rigaud2016toward}, the challenges of speech text recognition for comics are outlined, and several approaches for addressing these challenges are implemented. The study concludes that typewritten-like text is better recognized by generic OCR systems, while specifically trained OCR models can perform better for skewed, uppercase, and cursive fonts. However, our study shows that specifically trained models can outperform generic OCR systems, demonstrating the progress that has been made in the field since the previous study.

In \cite{textdetectionmanga_gobbo2020unconstrained}, pixel-level text annotations were curated for unconstrained text detection in the manga, and special evaluation metrics were developed to measure performance. In a more recent study, the Comic Onomatopoeia Dataset (COO) \cite{onomatopoeia_baek2022coo} presents a dataset of onomatopoeia (textual representations of sounds, states, or objects) for text detection, text recognition, and link prediction tasks. COO provides onomatopoeia in Japanese, pushing the limits of current irregular text detection and text recognition models. While our study focuses on providing an extensive dataset for comics, COO focuses more on the performance of models for their specific tasks. However, combining our research on both datasets can provide a more comprehensive understanding of the text modality of comics.

In \cite{ocrpipeline_hartel2021ocr}, Hartel and Dunst constructed an OCR pipeline for comics that begins with segmenting comics page components, such as panels and speech bubbles, and then extracts text from these components. They also analyzed the textual properties of their OCR results in terms of genre affiliation and page length. Our work differs from theirs because we use state-of-the-art deep learning-based text detection and text recognition models for OCR rather than the open-source software Calamari \cite{wick_calamari_2020}. Additionally, we make our models, training datasets, and OCR results for more than 2 million textboxes publicly available, whereas their data cannot be shared due to copyright restrictions.

\subsection{Datasets}
\label{subsection:datasets}

\begin{table}[hb!]
    \caption{ Overview of datasets in the domain of the comic.}
    \centering
    \resizebox{\textwidth}{!}{
        \begin{tabular}{@{}ccccc@{}}
            \toprule
            \multicolumn{1}{P{4cm}}{\centering \textbf{Dataset Name}}                                                             & \multicolumn{1}{P{3cm}}{\centering \textbf{Comic Book Release Dates}} & \multicolumn{1}{P{3cm}}{\centering \textbf{Top Level Item Count}} & \multicolumn{1}{P{3cm}}{\centering \textbf{Annotation Types}}                               & \multicolumn{1}{P{3cm}}{\centering \textbf{Public Availability}} \\ \midrule
            \multicolumn{1}{P{4cm}}{The Graphic Narrative Corpus (GNC) \cite{gncdataset_dunst2017graphic}}                        & 1975 - 2018                                                           & 240 Comic Books                                                   & \multicolumn{1}{P{3cm}}{panel, balloon, caption, diegetic text, text, character, object}    & NOT AVAILABLE                                                    \\ \arrayrulecolor{gray}\hline
            \multicolumn{1}{P{4cm}}{COMICS \cite{iyyer2017amazing}}                                                               & 1938 - 1956                                                           & 3948 Comic Books                                                  & \multicolumn{1}{P{3cm}}{panel, balloon, captions, text}                                     & AVAILABLE                                                        \\ \arrayrulecolor{gray}\hline
            \multicolumn{1}{P{4cm}}{Manga109 \cite{manga_109}}                                                                    & 1970 - 2010                                                           & 109 Mangas                                                        & \multicolumn{1}{P{3cm}}{panel, balloon, caption, onomatopoeia, text, character, face, body} & UPON REQUEST                                                     \\ \arrayrulecolor{gray}\hline
            \multicolumn{1}{P{4cm}}{DCM772 \cite{dcm772dataset_digital_comic_indexing}}                                           & 1938 - 1956                                                           & 27 Comic Books                                                    & \multicolumn{1}{P{3cm}}{panel, balloon, caption, character, face, body, type of character}  & AVAILABLE                                                        \\ \arrayrulecolor{gray}\hline
            \multicolumn{1}{P{4cm}}{Comic Onomatopoeia(COO) \cite{onomatopoeia_baek2022coo}}                                      & 1970 - 2010                                                           & 109 Mangas                                                        & \multicolumn{1}{P{3cm}}{onomatopoeia, text}                                                 & UPON REQUEST                                                     \\ \arrayrulecolor{gray}\hline
            \multicolumn{1}{P{4cm}}{eBDtheque \cite{eBDtheque2013}}                                                               & 1905 - 2012                                                           & 100 Pages                                                         & \multicolumn{1}{P{3cm}}{panel, balloon, text}                                               & UPON REQUEST                                                     \\ \arrayrulecolor{gray}\hline
            \multicolumn{1}{P{4cm}}{Multimodal Emotion Recognition on Comics scenes (EmoRecCom) \cite{emoreccom_nguyen2021icdar}} & 1938 - 1956                                                           & 8258 Panels                                                       & \multicolumn{1}{P{3cm}}{multiple emotion labels for panels}                                 & AVAILABLE                                                        \\ \bottomrule
        \end{tabular}}
    \label{table:datasets_overview}
\end{table}

Although there are not many datasets available for the comic domain due to copyright issues, a few datasets have been made publicly available. In Table \ref{table:datasets_overview}, you can see a complete list of datasets focused on comic books and manga within our knowledge.

The Graphic Narrative Corpus (GNC) \cite{gncdataset_dunst2017graphic} is a unique dataset to serve both computational sciences and social sciences. The dataset provides a variety of metadata, such as genre, author, and geographic origins, along with annotated data to the extent of eye gaze data. However, Unfortunately, the dataset can not be accessed publicly.

COMICS \cite{iyyer2017amazing} is the dataset with the most comic books and is the dataset we use for our study. They also propose downstream tasks for the domain of the comic. These are text cloze, visual cloze, and character coherence.

The Manga109 \cite{manga_109} dataset contains 109 volumes of manga by 93 different authors. These volumes were released from the 1970s to the 2010s and fall into 12 different genres, such as fantasy, humor, and sports.

DCM772 \cite{dcm772dataset_digital_comic_indexing} is one of the few publicly available datasets because of how it was collected. The comic books in the dataset come from a free public domain collection called the Digital Comic Museum \footnote{\url{http://digitalcomicmuseum.com}}. This dataset annotates different character types, such as animal-like or human-like.

Comic Onomatopoeia (COO) \cite{onomatopoeia_baek2022coo} is a dataset that extends the Manga109 dataset with onomatopoeia annotations. Because COO focuses on onomatopoeia, it provides a challenging text detection and recognition dataset with various shapes, styles, and extreme curves. They also introduce a novel task of linking truncated texts.

The eBDtheque dataset \cite{eBDtheque2013} is a pioneering dataset for comic books, including American comics, mangas, and Franco-Belgian comic book albums. It provides panels, balloons, text lines, and pages.

The Multimodal Emotion Recognition on Comics scenes (EmoRecCom) \cite{emoreccom_nguyen2021icdar} dataset is a small subset of the COMICS dataset with labels for eight different emotions for each scene. Our COMICS Text+ dataset may be able to improve the text quality of this dataset, as its panels are sourced from the COMICS dataset.

\section{Methodology}
\label{section:methodology}

\begin{figure}[hb!]
    \centering
    \includegraphics[width=1.0\textwidth]{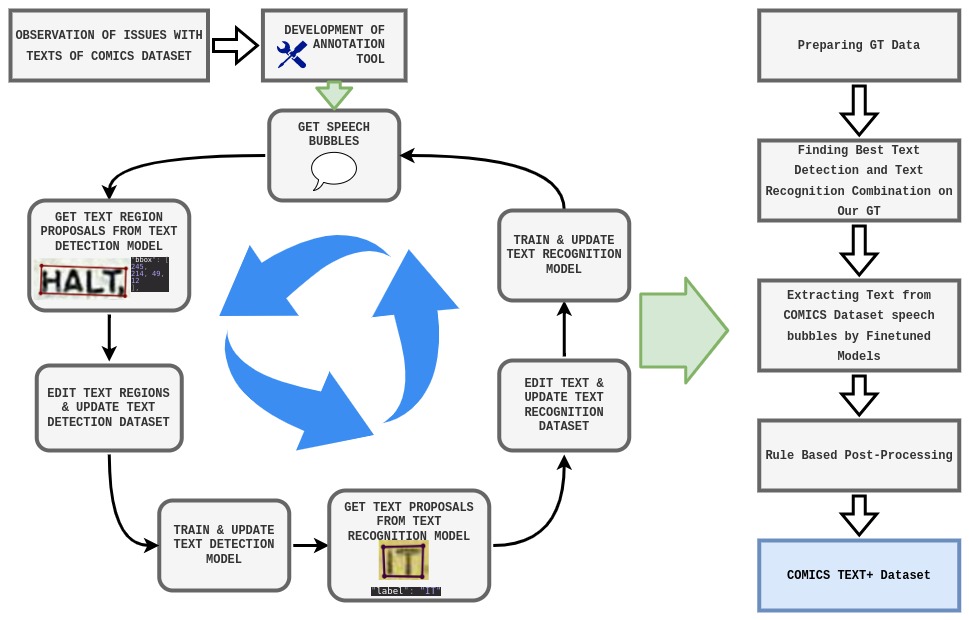}
    \caption{The complete process and pipeline of our approach and dataset.}
    \label{fig:pipeline_process}
\end{figure}

Our approach to improving the text data in the COMICS dataset \cite{iyyer2017amazing} consists of three main stages, as shown by the green arrows in Figure \ref{fig:pipeline_process}. In the first stage, we identified the shortcomings and issues with the text data in the COMICS dataset and developed an annotation tool by forking the LabelMe tool \cite{wada2018labelme}. Our motivation for doing this is that the previous works that proposed the COMICS dataset did not provide any metrics \cite{iyyer2017amazing,iyyer2017discourse}.

In the second stage, we used the panels and speech bubbles cropped out of the pages of comics in the COMICS dataset to initiate the machine learning lifecycle. During this stage, we annotated the data and trained out-of-the-box models successively until the model performance was stable. At the end of this stage, we created text detection and text recognition datasets using more than 1000 speech bubbles (see Table \ref{table:final_det_recog_dataset_size}).

In the third stage, we used our text detection and text recognition datasets to train text detection and text recognition models provided in the MMOCR framework \cite{mmocr2021}. We then benchmarked the performance of these models on the dataset. We selected the top-performing model combination for detection and recognition and used it to measure the peak performance on a carefully selected set of speech bubbles, referred to as the ground truth (GT). Finally, we used the selected model combination to run over the entire set of speech bubbles in the COMICS dataset and post-processed the resulting data to create the COMICS Text+ dataset.

\subsection{Textual Flaws of COMICS Dataset}
\label{subsection:comics_flaws}

For the COMICS dataset, it is stated in \cite{iyyer2017amazing} and \cite{iyyer2017discourse} that the OCR data has detection and recognition flaws  which were attempted to be mitigated through post-processing. Those flaws are as follows:

\begin{itemize}
    \item Detection of short words
    \item Detection of punctuation marks
    \item Recognition of the first letter of the word
\end{itemize}

In addition to those cases, we realized more issues with the dataset. Such as:

\begin{itemize}
    \item Skipping lines: Some lines of text were not detected by the OCR system.
    \item Total misrecognition of words: In some cases, entire words were misrecognized by the OCR system.
    \item Difficulty recognizing certain fonts: The OCR system had difficulty recognizing certain fonts, such as italic fonts.
\end{itemize}

Examples of these errors can be seen in Table \ref{table:comics_dataset_ocr_flaws}. The quantitative evaluation results of the COMICS dataset OCR compared to our ground truth are shown in Table \ref{table:final_results_comics_vs_ours}, where it can be seen that the character recall of the COMICS dataset OCR is lower than our final results, supporting our qualitative analysis.

\begin{table}[htbp!]
    \caption{Results of qualitative error analysis on COMICS Dataset OCR data.}
    \centering
    \begin{footnotesize}
        \begin{tabular}{ | c | m{5cm} | m{4cm} | }
            \hline
            \begin{normalsize}
                \textbf{Speech Bubble}
            \end{normalsize}
             &
            \begin{normalsize}
                \textbf{COMICS Dataset OCR}
            \end{normalsize}
             &
            \begin{normalsize}
                \textbf{Ground Truth}
            \end{normalsize}
            \\ \hline
            \begin{minipage}{.3\textwidth}
                \includegraphics[width=\linewidth]{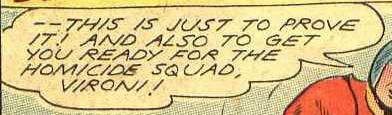}
            \end{minipage}
             &
            / 7 ./ anoa < s 70 get you ready for 7 \% e - vro // / yo
             &
            - - this is just to prove it ! and also to get you ready for the homicide squad , vironi !
            \\
            \hline

            \begin{minipage}{.3\textwidth}
                \includegraphics[width=0.5\linewidth]{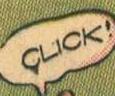}
            \end{minipage}
             &
            ok ,
             &
            click !
            \\
            \hline

            \begin{minipage}{.3\textwidth}
                \includegraphics[width=\linewidth]{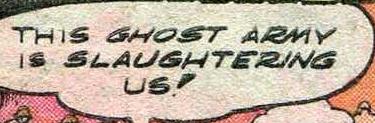}
            \end{minipage}
             &
            is slaughter ,
             &
            this ghost army is slaughtering us !
            \\ \hline

            \begin{minipage}{.3\textwidth}
                \includegraphics[width=\linewidth]{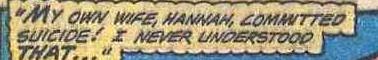}
            \end{minipage}
             &
            my oww wae , mamma , comantted
             &
            " my own wife , hannah , committed suicide ! i never understood that . . . "
            \\
            \hline

            \begin{minipage}{.3\textwidth}
                \includegraphics[width=\linewidth]{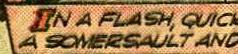}
            \end{minipage}
             &
            a sameersalaltano
             &
            in a flash , quick ! a somersault and
            \\
            \hline

            \begin{minipage}{.3\textwidth}
                \includegraphics[width=0.5\linewidth]{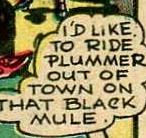}
            \end{minipage}
             &
            that black id ride like of on to plummer out mule
             &
            i ' d like to ride plummer out of town on that black mule
            \\

            \hline
        \end{tabular}
    \end{footnotesize}
    \label{table:comics_dataset_ocr_flaws}
\end{table}

\subsection{Annotation Tool \& Creation of New Comics Text Detection \& Text Recognition Datasets  }

To improve the text data in the COMICS dataset, we developed an annotation tool based on the LabelMe tool \cite{wada2018labelme}. In addition to the features of the original tool, we added two new modes, "Detect Text" and "Detect and Recognize Text", which use inference from text detection and text recognition models to speed up the annotation process. As we continued to fine-tune our models with more data, this functionality helped to reduce the time required for annotation.

This tool is essential to our approach, as it allows us to create new text detection and text recognition datasets and continuously train text detection and text recognition models. The cyclic process of using the tool is as follows:

\begin{itemize}
    \item Randomly crop speech bubbles or panels from the COMICS dataset, selecting at least a couple of images from each comic book.
    \item Use a pretrained model to generate text region proposals for the selected images and annotate them for text detection.
    \item Edit the proposed text regions as needed.
    \item Train a text detection model (we used the DBNet++ model \cite{DBNETPP_liao2022real}) and update it for use in the text recognition pipeline.
    \item Use the updated text detection model along with the pretrained text recognition model to generate text region and text proposals for the selected images, and annotate them for text recognition.
    \item Edit the proposed text regions and texts as needed.
    \item Train a text recognition model (we used the NRTR model \cite{sheng2019nrtr}) and update it for use in the text recognition pipeline.
    \item Repeat the process for additional images to continue improving the performance of the models.
\end{itemize}

After each training cycle, we evaluate our models' performance using the corresponding training sets and ground truth data. This allows us to monitor the accuracy of our annotations and make any necessary adjustments to our annotation policy (see Subsection \ref{subsection:comics_text_detection_recognition_datasets}). By regularly assessing the performance of our models, we can ensure that our annotation process is effective and that our models are learning from the data as intended.

\subsubsection{Detection Mode}

\begin{figure}[htbp!]
    \centering
    \includegraphics[width=\textwidth]{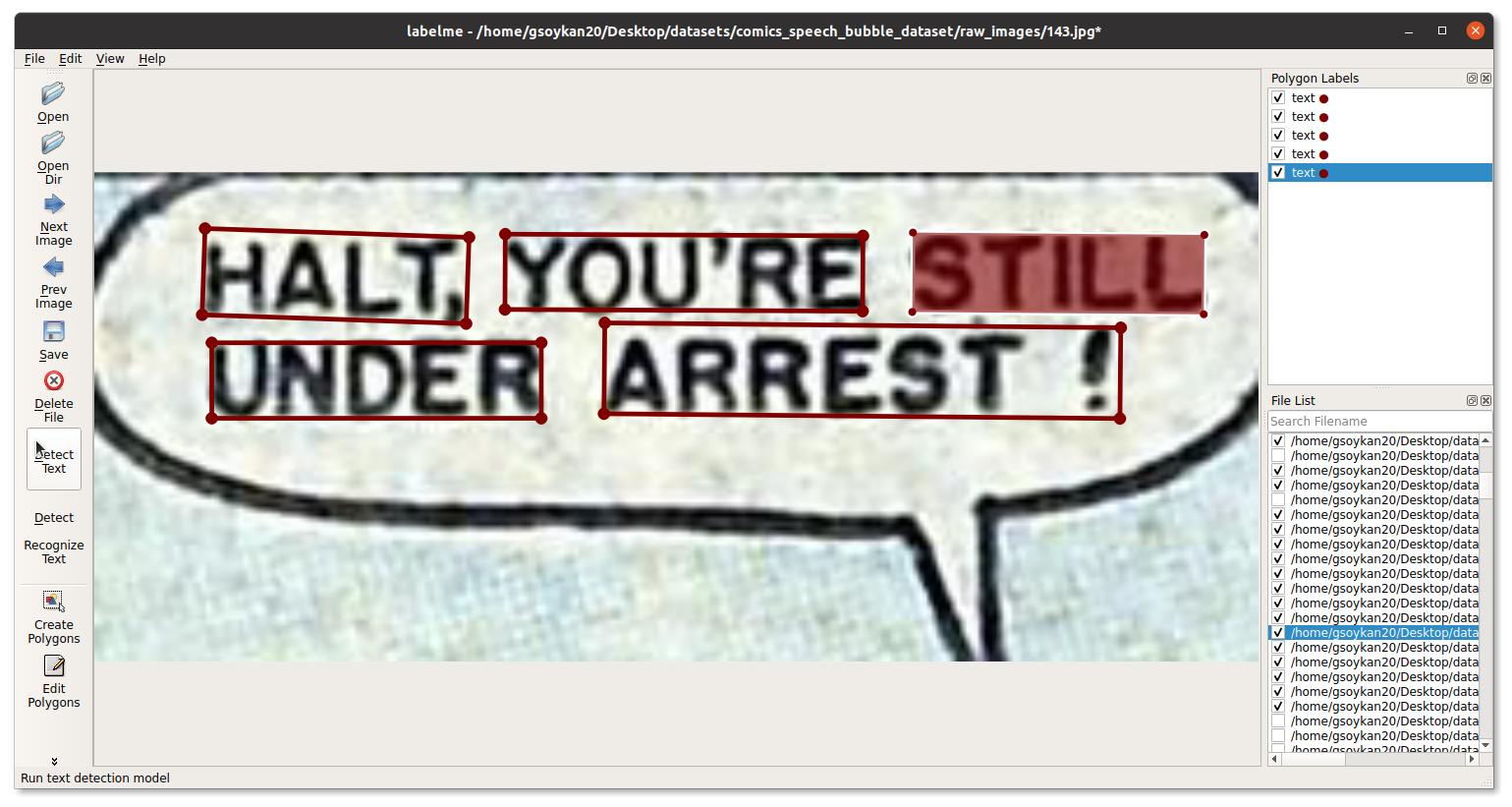}
    \caption{Text detection mode of our annotation tool. Text regions, mostly words,  can be estimated by a backbone text detection model and edited.}
    \label{fig:annot_tool_detection}
\end{figure}

In detection mode, a text detection model is used to infer the polygons of the text regions in an image. The user can then edit, add, or remove these polygons to create finely detailed annotations. This is the approach we took when creating the \textit{COMICS Text+: Detection}. By allowing for manual adjustment of the detected polygons, we were able to create a dataset with high-quality annotations that are well-suited for training text detection models.

\subsubsection{Recognition Mode}

\begin{figure}[htbp!]
    \centering
    \includegraphics[width=\textwidth]{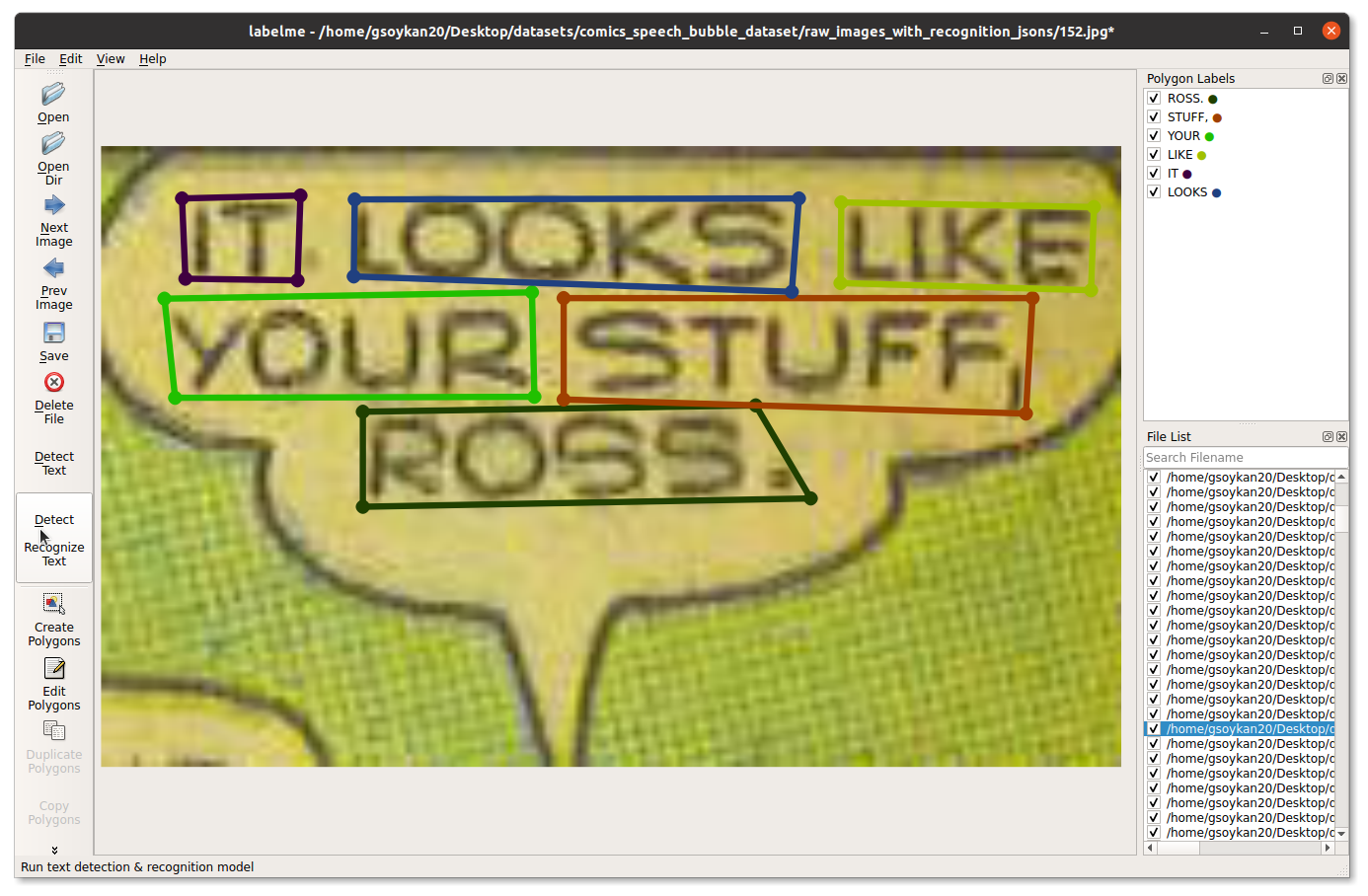}
    \caption{Text recognition mode of our annotation tool. Recognition results can be estimated by a backbone text recognition model and edited.}
    \label{fig:annot_tool_recognition}
\end{figure}

In recognition mode, a pipeline consisting of a text detection model and a text recognition model is used to infer the polygons of the text regions in an image and the corresponding text within those regions. The user can then edit, add, or remove these annotations to create finely detailed ones. This is the approach we took when creating the \textit{COMICS Text+: Recognition}. Using a combination of text detection and recognition models, we were able to generate initial annotations that were then refined by manual editing, resulting in a dataset with high-quality annotations suitable for training text recognition models.

\subsubsection{Conversion to Dataset Format}

Once all the annotations are complete, we can export those in dataset formats with the conversions scripts in our pipeline, shared with this \href{https://github.com/gsoykan/labelme}{repository}.  For the detection dataset, we use the \textit{IcdarDataset} \cite{karatzas2015icdar} \footnote{
    \url{https://mmocr.readthedocs.io/en/latest/tutorials/dataset_types.html\#icdardataset
    }} format and for the recognition dataset, we use the \textit{OCRDataset} \footnote{
    \url{https://mmocr.readthedocs.io/en/latest/tutorials/dataset_types.html\#ocrdataset
    }} format. Using these standardized formats, we can easily integrate our datasets into existing pipelines for training and evaluating text detection and recognition models.

\subsubsection{Creation of COMICS Text+: Detection \& Recognition Datasets}
\label{subsection:comics_text_detection_recognition_datasets}

The \textit{COMICS Text+: Detection} and \textit{COMICS Text+: Recognition} datasets were created using our annotation tool and pipeline. The text detection dataset consists of annotated text regions in speech bubbles or panels. Because of the nature of the domain, these text regions are mostly individual words. However, defining the boundaries of text regions can be challenging in this domain. Our annotation policy was refined through several iterations. Initially, we annotated text regions with large margins and included punctuation marks within a text region that contained a word. This led to models trained on this data producing erroneous predictions of text regions that were often intersecting. Our final policy for text detection annotations involves using a low margin for words and excluding punctuation marks if they are not close enough to the words (i.e., their distance from the words must be no more than twice the letter spacing). Another factor that complicates annotations is comic books' varying spacing between words and hyphens. We follow the same strategy with hyphens as we do with punctuation marks.

For the text recognition dataset, we annotate text regions with their corresponding text. In this case, the focus is on recognizing the text within the text regions, so we carefully select regions with clear text. The challenges of text detection do not apply in this context. Another difference between the two datasets is that we sometimes skip some text regions within an image because of their irregular text shapes or poor quality (e.g., if the text is obscured or worn). Since we cannot skip these text regions in the text detection dataset, the text detection and text recognition datasets may have different images and annotations. The final statistics of our text detection and text recognition datasets are shown in Table \ref{table:final_det_recog_dataset_size}.

\begin{table}
    \caption{Statistics describing splits of dataset sizes constituting \textit{COMICS Text+: Detection \& Recognition} datasets. The dataset is used to train and evaluate text detection and recognition models. In our experiments, the training set size can be different, but the test split is always used as it is. }
    \centering
    \resizebox{\textwidth}{!}{
        \begin{tabular}{@{}ccccccc@{}}
            \toprule
                                      & \multicolumn{2}{l}{\textbf{Training}} & \multicolumn{2}{l}{\textbf{Validation}} & \multicolumn{2}{l}{\textbf{Test}}                                               \\ \midrule
                                      & \# Images                             & \# Annotations                          & \# Images                         & \# Annotations & \# Images & \# Annotations \\
            \textbf{Text Detection}   & 1012                                  & 18494                                   & 50                                & 790            & 50        & 986            \\
            \textbf{Text Recognition} & 906                                   & 15635                                   & 50                                & 759            & 50        & 714            \\ \bottomrule
        \end{tabular}}
    \label{table:final_det_recog_dataset_size}
\end{table}

\subsection{Creating Text Ground Truth Data From COMICS Dataset}
\label{subsection:creating_gt}

To evaluate the full performance of fine-tuned text detection and text recognition models, ground truth (GT) data is needed. Since no such data was available for Golden Age comics, we created and organized our own GT data. This data consists of tuples of speech bubbles or narrative boxes and their corresponding text. Unlike the text detection and recognition datasets, the GT data does not include labels for individual text regions; instead, it covers all the text in an image. We carefully selected the speech bubbles for the GT data to ensure high-quality annotations. This GT data allows us to thoroughly evaluate the performance of our models on the COMICS Text+ datasets.

\begin{figure}[htbp!]
    \centering
    \includegraphics[width=0.25\textwidth]{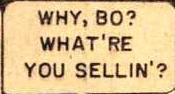}
    \includegraphics[width=0.25\textwidth]{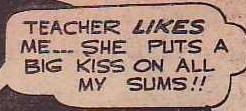}
    \includegraphics[width=0.25\textwidth]{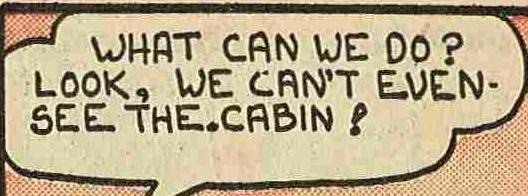}
    \includegraphics[width=0.25\textwidth]{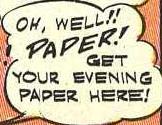}
    \includegraphics[width=0.25\textwidth]{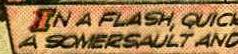}
    \caption{Five examples from selected speech bubbles for ground truth data.}
    \label{fig:gt_selected}
\end{figure}

Our policy for selecting speech bubbles for the GT data is to ensure that all the text in the image is clearly visible and readable. The selected images come from different comic books and have various artistic styles. In some cases, there may be residual characters on the margins of the speech bubbles, but this is acceptable as long as the characters are not fully visible. This allows us to see how the text detector performs in the presence of residual text and ensures that it does not incorrectly label these as text regions. However, we do not include examples with half-words or full characters in the GT data because this would introduce irregular text and speech bubble detection problems that would skew the evaluation of our models' performance. Figure \ref{fig:gt_selected} shows some examples of our selected GT data.

\begin{figure}[htbp!]
    \centering
    \includegraphics[width=0.2\textwidth]{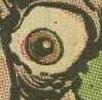}
    \includegraphics[width=0.2\textwidth]{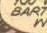}
    \includegraphics[width=0.2\textwidth]{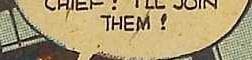}
    \includegraphics[width=0.2\textwidth]{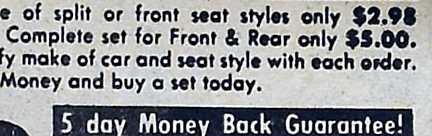}
    \includegraphics[width=0.2\textwidth]{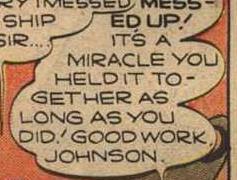}
    \caption{Five examples of speech bubbles that are not eligible for ground truth data.}
    \label{fig:gt_unselected}
\end{figure}

To create our GT data, we randomly cropped multiple images from different comic series using the coordinates of the text boxes provided in the COMICS dataset. We then filtered this data to refine our GT. Figure \ref{fig:gt_unselected} shows some examples of the data that was filtered out. We did not include images that had one or more of the following features:

\begin{itemize}
    \item Partial words or texts
    \item Parts of advertisement pages
    \item No text
    \item Multiple panels
    \item Irregular or specialized fonts
\end{itemize}

By applying these filters, we were able to create a GT dataset with high-quality annotations that accurately reflect the text in the images. This allows us to accurately evaluate the performance of our text detection and recognition models.

\subsection{Benchmarking the Detection and Recognition Models}
\label{subsection:best_detection_recognition_models}

We trained fourteen text detection, and ten text recognition models using the MMOCR toolkit \cite{mmocr2021} and benchmarked their performance. After finding the top-performing models on the test split, we evaluated the combination of top three models on the GT data to determine which combination of models gives us the best results. This allows us to identify the most effective models for detecting and recognizing text in Golden Age comic images.

\subsubsection{Detection Model Benchmark}

We benchmarked fourteen recently published, high-performing pre-trained text detection models on our \textit{COMICS Text+: Detection} dataset. The results are shown in Table \ref{table:detector_benchmark}. The models we used are as follows:

\begin{itemize}
    \item \textbf{\currier{DB\_r18}:} DBNet \cite{DBNET_Liao_Wan_Yao_Chen_Bai_2020} with ResNet-18 \cite{resnet} backbone pretrained on ICDAR2015 \cite{karatzas2015icdar} dataset
    \item \textbf{\currier{DB\_r50}:} DBNet with ResNet-50 backbone pretrained on ICDAR2015 dataset
    \item \textbf{\currier{DBPP\_r50}:} DBNet++ \cite{DBNETPP_liao2022real} with ResNet-50 \cite{resnet} backbone pretrained on ICDAR2015 dataset
    \item \textbf{\currier{DRRG}:} Deep Relational Reasoning Graph Network \cite{DRRG_zhang2020drrg} pretrained on CTW-1500 dataset \cite{ctw1500-yuliang2017detecting}
    \item \textbf{\currier{FCE\_IC15}:} FCENet \cite{FCENET_zhu2021fourier} with ResNet-50 backbone pretrained on ICDAR2015 dataset
    \item \textbf{\currier{FCE\_CTW\_DCNv2}:} FCENet \cite{FCENET_zhu2021fourier} with ResNet-50 and deformable convolutional networks (DCN) \cite{dai2017deformable} backbone pretrained on CTW-1500 dataset
    \item \textbf{\currier{MaskRCNN\_IC17 }:} Mask R-CNN \cite{MASK_RCNN_8237584} pretrained on ICDAR2017 \cite{ICDAR2017-8270165} dataset
    \item \textbf{\currier{MaskRCNN\_IC15}:} Mask R-CNN pretrained on ICDAR2015
    \item \textbf{\currier{MaskRCNN\_CTW}:} Mask R-CNN pretrained on CTW-1500 dataset
    \item \textbf{\currier{PANet\_CTW}:} Pixel Aggregation Network (PAN) \cite{PANET_WangXSZWLYS19} pretrained on CTW-1500 dataset
    \item \textbf{\currier{PANet\_IC15}:} PAN pretrained on ICDAR2015
    \item \textbf{\currier{PS\_IC15}:} Progressive Scale Expansion Network (PSENet) \cite{PSENET_wang2019shape} pretrained on ICDAR2015
    \item \textbf{\currier{PS\_CTW}:} PSENet pretrained on CTW-1500 dataset
    \item \textbf{\currier{TextSnake}:} TextSnake \cite{TEXTSNAKE_long2018textsnake} pretrained on CTW-1500 dataset
\end{itemize}

\subsubsection{Training Details}

During training, we used a single NVIDIA V100 GPU. The batch sizes for each training were varied to maximize the use of the GPU. We used the Adam optimizer with a learning rate of 0.0001 and applied gradient clipping with a max norm value of 0.5. We also used a step learning rate scheduler at the third and fourth epochs, reducing the learning rate by 0.1. The models were trained for six epochs, and the best-performing model in each case was selected for evaluation on the test set based on the hmean metric. All experiments were configured using the MMOCR \cite{mmocr2021} training scripts.

\begin{table}
    \caption{Best text detection model results trained on our dataset sorted by hmean. }
    \centering
    \begin{tabular}{| l | r | r | r |}
        \hline
        \textbf{Text Detection Model}    & \textbf{Recall} & \textbf{Precision} & \textbf{Hmean} \\
        \hline

        \rowcolor{top_1} DBPP\_r50       & 0.933063        & 0.971489           & 0.951888       \\
        \rowcolor{top_2} FCE\_CTW\_DCNv2 & 0.929006        & 0.967265           & 0.94775        \\
        \rowcolor{top_3} MaskRCNN\_IC17  & 0.924949        & 0.968153           & 0.946058       \\
        PS\_IC15                         & 0.931034        & 0.960251           & 0.945417       \\
        MaskRCNN\_CTW                    & 0.938134        & 0.943878           & 0.940997       \\
        MaskRCNN\_IC15                   & 0.925963        & 0.944157           & 0.934972       \\
        DB\_r50                          & 0.910751        & 0.945263           & 0.927686       \\
        PS\_CTW                          & 0.915822        & 0.938669           & 0.927105       \\
        DB\_r18                          & 0.896552        & 0.954644           & 0.924686       \\
        PANet\_IC15                      & 0.902637        & 0.938819           & 0.920372       \\
        TextSnake                        & 0.859026        & 0.948488           & 0.901543       \\
        FCE\_IC15                        & 0.901623        & 0.893467           & 0.897527       \\
        DRRG                             & 0.850913        & 0.946953           & 0.896368       \\
        PANet\_CTW                       & 0.841785        & 0.926339           & 0.88204        \\
        \hline
    \end{tabular}
    \label{table:detector_benchmark}
\end{table}

\subsection{Recognition Model Benchmark}

We benchmark recently published and performant fourteen pre-trained text recognition models on our \textit{COMICS Text+: Recognition} dataset. The results are shown in Table \ref{table:recognition_benchmark}. The models we used are as follows:

\begin{itemize}
    \item \textbf{\currier{ABINet}:} ABINet \cite{ABINet-fang2021read} pretrained on Synth90k \cite{Syn90k-Jaderberg16} and SynthText \cite{SynthText-Gupta16}
    \item \textbf{\currier{CRNN}:} Convolutional Recurrent Neural Network (CRNN) \cite{CRNN-shi2016end} pretrained on Synth90k
    \item \textbf{\currier{MASTER}:} MASTER \cite{Lu2021MASTER} pretrained on Synth90k, SynthText, and SynthAdd \cite{SAR-li2019show}
    \item \textbf{\currier{NRTR\_1/8-1/4}:} NRTR \cite{sheng2019nrtr} with  the height of feature from the backbone is 1/16 of the input image, where 1/8 for width and pretrained on Synth90k and SynthText
    \item \textbf{\currier{NRTR\_1/16-1/8}:} NRTR with the height of feature from the backbone is 1/8 of the input image, where 1/4 for width and pretrained on Synth90k and SynthText
    \item \textbf{\currier{RobustScanner}:} RobustScanner \cite{yue2020robustscanner} pretrained on multiple ICDAR datasets, Synth90k, SynthText, SynthAdd and more \footnote{
              \url{https://mmocr.readthedocs.io/en/latest/textrecog_models.html\#robustscanner
              }}
    \item \textbf{\currier{SAR}:} Show-Attend-and-Read (SAR) \cite{SAR-li2019show} pretrained on multiple ICDAR datasets, Synth90k, SynthText, SynthAdd and more \footnote{
              \url{https://mmocr.readthedocs.io/en/latest/textrecog_models.html\#sar
              }}
    \item \textbf{\currier{SATRN}:} Self-Attention Text Recognition Network (SATRN) \cite{SATRN} pretrained on Synth90k and SynthText
    \item \textbf{\currier{SATRN\_sm}:} SATRN with a reduced number of encoder and decoder layers pretrained on Synth90k and SynthText
    \item \textbf{\currier{CRNN-TPS}:} CRNN-STN \cite{CRNNTPS-shi2016robust} pretrained on Synth90k
\end{itemize}

\paragraph{Training Details}

During training, we used a single NVIDIA V100 GPU. The batch sizes for each training were varied to maximize the use of the GPU. We used the Adam optimizer with a learning rate of 0.0001, and applied gradient clipping with a max norm value of 0.5. We also used a step learning rate scheduler at the third and fourth epochs, reducing the learning rate by 0.1. The models were trained for six epochs, and the best-performing model in each case was selected for evaluation on the test set based on the 1-N.E.D. metric. All experiments were configured using the MMOCR \cite{mmocr2021} training scripts.

\begin{table}
    \caption{Best text recognition model results trained on our dataset sorted by 1-N.E.D. }
    \centering
    \resizebox{\textwidth}{!}{
        \begin{tabular}{| l | r | r | r | r | r |}
            \hline
            \multicolumn{1}{|p{4cm}|}{\centering \textbf{Text Recognition Model}} & \multicolumn{1}{|p{2cm}|}{\centering \textbf{Char. Recall}} & \multicolumn{1}{|p{2cm}|}{\centering \textbf{Char. Precision}} & \multicolumn{1}{|p{2cm}|}{\centering \textbf{Word Acc.}} & \multicolumn{1}{|p{3.2cm}|}{\centering \textbf{Word Acc. Ignore Symbol}} & \multicolumn{1}{|p{2cm}|}{\centering \textbf{1-N.E.D}} \\
            \hline
            \rowcolor{top_1} MASTER                                               & 0.9962                                                      & 0.9954                                                         & 0.9593                                                   & 0.9832                                                                   & 0.9923                                                 \\
            \rowcolor{top_2} NRTR\_1/8-1/4                                        & 0.995                                                       & 0.9935                                                         & 0.9579                                                   & 0.9804                                                                   & 0.9919                                                 \\
            \rowcolor{top_3} NRTR\_1/16-1/8                                       & 0.9935                                                      & 0.9924                                                         & 0.9509                                                   & 0.9748                                                                   & 0.9918                                                 \\
            RobustScanner                                                         & 0.9924                                                      & 0.9924                                                         & 0.9425                                                   & 0.9705                                                                   & 0.9855                                                 \\
            SAR                                                                   & 0.9928                                                      & 0.9931                                                         & 0.9453                                                   & 0.9705                                                                   & 0.9825                                                 \\
            SATRN                                                                 & 0.9909                                                      & 0.992                                                          & 0.9425                                                   & 0.9649                                                                   & 0.9811                                                 \\
            SATRN\_sm                                                             & 0.9867                                                      & 0.9882                                                         & 0.9299                                                   & 0.9495                                                                   & 0.9766                                                 \\
            ABINet                                                                & 0.995                                                       & 0.7867                                                         & 0.7237                                                   & 0.7293                                                                   & 0.8449                                                 \\
            CRNN-TPS                                                              & 0.9863                                                      & 0.785                                                          & 0.7083                                                   & 0.7153                                                                   & 0.8411                                                 \\
            CRNN                                                                  & 0.984                                                       & 0.7843                                                         & 0.6971                                                   & 0.7069                                                                   & 0.8397                                                 \\
            \hline
        \end{tabular}}
    \label{table:recognition_benchmark}
\end{table}

\subsubsection{Finding Best Text Detection and Text Recognition Models for OCR}

Our text detection and text recognition benchmarks showed that we have close-performing models after training on our text detection and text recognition datasets. Even though their results are close to each other in metrics, we realized their failure points differ. Because of this reason, we decided to measure the OCR performance of the top three of their combinations. Even though we chose DBNET++ in our annotation - training - evaluation cycle due to qualitative evaluation, after this process, we saw that the FCENET text detection backbone works best by a high margin for the OCR metrics in the domain of the comics (see Table \ref{table:det_recog_combo}). From our results, we can clearly say that for the current OCR of comics, the deciding factor is the text detection model used. The OCR performance does not vary by high values when we keep the same text detection model and change the text recognition model. However, the OCR results change dramatically when keeping the text recognition model the same and changing the text detection model.

\begin{table}
    \caption{Results of top-performing text detection and text recognition models trained on our dataset measured with respect to ground truth data. }
    \resizebox{\textwidth}{!}{
        \begin{tabular}{@{}ccccccc@{}}
            \toprule
            \multicolumn{1}{p{3.2cm}}{\centering Recognition / Detection Models} & \multicolumn{2}{c}{\textbf{NRTR\_1/16-1/8}}                   & \multicolumn{2}{c}{\textbf{NRTR\_1/8-1/4}} & \multicolumn{2}{c}{\textbf{MASTER}}                                                                                                                          \\ \midrule
                                                                                 & \multicolumn{1}{p{3.2cm}}{\centering Word Acc. Ignore Symbol} & 1 - N.E.D.                                 & \multicolumn{1}{p{3.2cm}}{\centering Word Acc. Ignore Symbol} & 1 - N.E.D. & \multicolumn{1}{p{3.2cm}}{\centering Word Acc. Ignore Symbol} & 1 - N.E.D.      \\
            \textbf{MaskRCNN\_IC17}                                              & 0.076                                                         & 0.7557                                     & 0.074                                                         & 0.7579     & 0.076                                                         & 7631            \\
            \textbf{FCE\_CTW\_DCNv2}                                             & 0.714                                                         & 0.9759                                     & 0.704                                                         & 0.9774     & \textbf{0.722}                                                & \textbf{0.9782} \\
            \textbf{DBPP\_r50}                                                   & 0.516                                                         & 0.9475                                     & 0.5932                                                        & 0.9615     & 0.6052                                                        & 0.9631          \\ \bottomrule
        \end{tabular}}

    \label{table:det_recog_combo}

\end{table}

\subsubsection{Text Detection and Text Recognition and OCR Performances by Data Size}

We investigated the performance of our most successful text detection and recognition models based on OCR results of GT data, as the number of training data varies. This allows us to determine the minimum amount of data needed to achieve optimal model performance. Since annotating and labeling data can be labor-intensive and time-consuming, we aimed to provide a baseline for data count in different comic domains and inform future studies on this topic. We focused on understanding the relationship between the number of training data and model performance in the context of text detection and recognition in comics.

\begin{figure}[htbp!]
    \centering
    \includegraphics[width=0.475\textwidth]{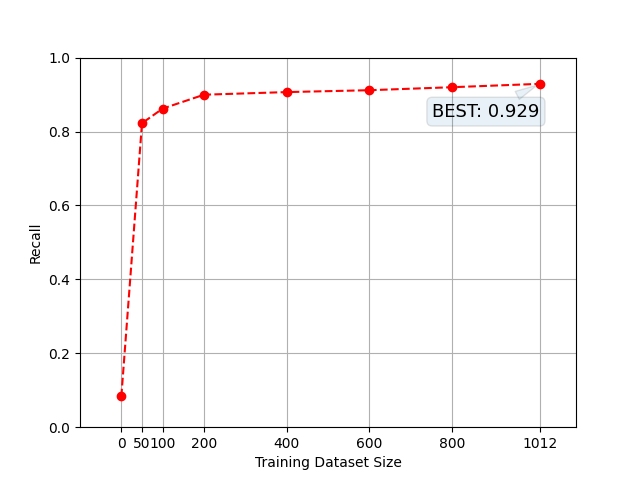}
    \includegraphics[width=0.475\textwidth]{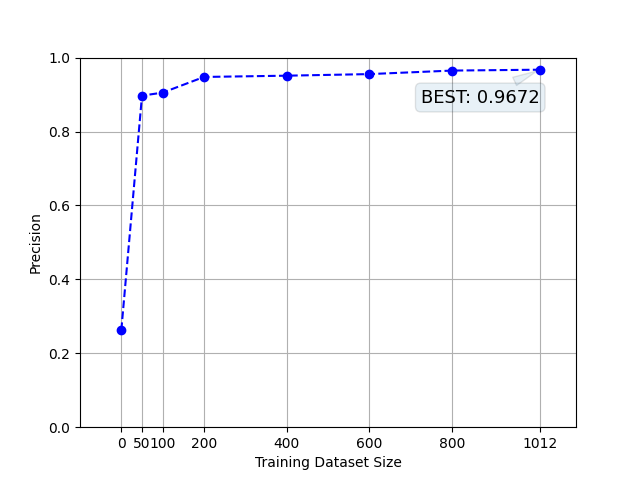}
    \includegraphics[width=0.5\textwidth]{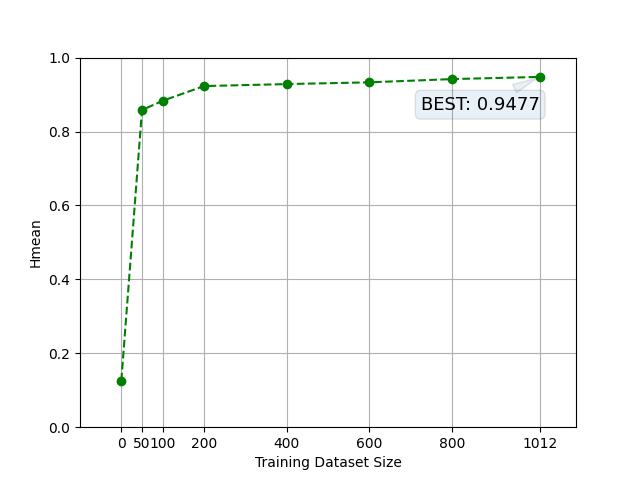}
    \caption{Recall, precision, and hmean graphs of the most performant detection backbone on the test dataset, FCE\_CTW\_DCNv2, for varying training dataset size.}
    \label{fig:best_det_data_size}
\end{figure}

For text detection we show the results in Figure \ref{fig:best_det_data_size} for precision, recall, and hmean. The interesting finding here is that since we already use a pretrained model, they adapt our domain with as low as fifty training samples and within a few epochs. Usually, it takes hundreds of epochs for the text detection model to learn from random initialization. After the training sample count is increased to two hundred, results do not change much, although as we increase the data, the model becomes more and more performant.

\begin{figure}[htbp!]
    \centering
    \includegraphics[width=0.475\textwidth]{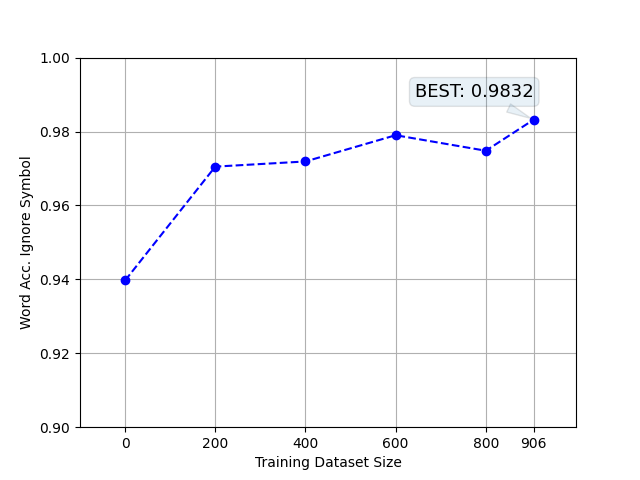}
    \includegraphics[width=0.475\textwidth]{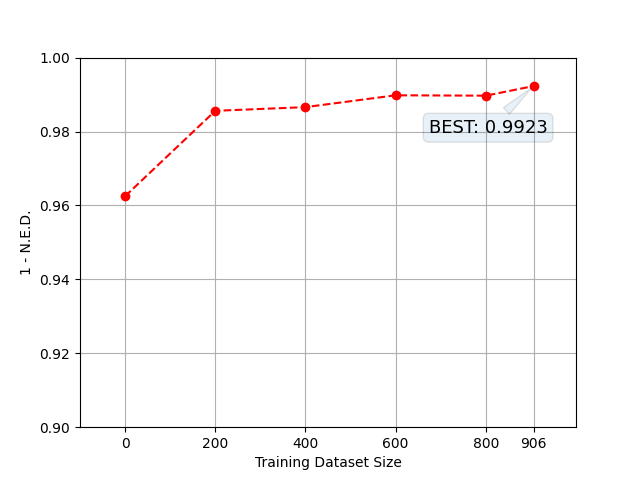}
    \caption{Word accuracy ignoring symbols, and 1 - N.E.D. graphs of the most performant recognition backbone on the test dataset, MASTER, for varying training dataset size.}
    \label{fig:best_recog_data_size}
\end{figure}

The same trend is observed in our text recognition model, as shown in Figure \ref{fig:best_recog_data_size}, where we measured word accuracy by ignoring symbols and using one minus normalized edit distance. However, the impact of domain adaptation on model performance is not as significant in this case, with only a two \% - three \% improvement in performance. In contrast, we see a much greater increase in performance for text detection, with an improvement of more than eight times.

\begin{figure}[htbp!]
    \centering
    \includegraphics[width=0.475\textwidth]{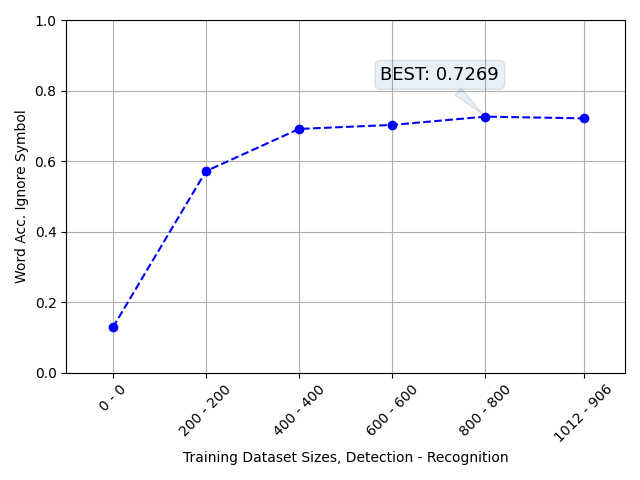}
    \includegraphics[width=0.475\textwidth]{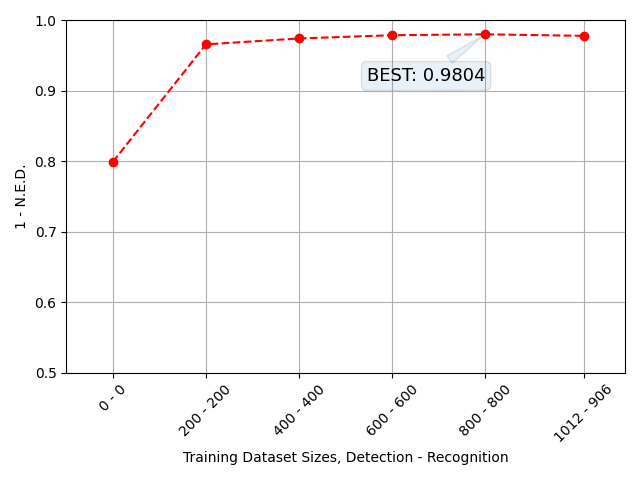}
    \caption{Word accuracy ignoring symbols, and 1 - N.E.D. graphs of the most performant detection and recognition model type, FCE\_CTW\_DCNv2 - MASTER, on the ground truth data that are trained on varying training dataset size combinations.}
    \label{fig:best_det_recog_data_size_gt}
\end{figure}

In our final step of performance measurement, we varied the dataset size and measured the performance of a combination of text detection and text recognition models (see Figure \ref{fig:best_det_recog_data_size_gt}). We used the one minus normalized edit distance metric to evaluate the model's performance and found that near-top performance was achieved with two hundred training sample sizes. However, when measuring word accuracy while ignoring symbols, we found that this level of performance was reached with four hundred training samples. This demonstrates the sensitivity of word accuracy to the model's performance, and suggests that four hundred training samples for both text detection and text recognition can be a good baseline for comics.

\section{Creating 'COMICS Text+ Dataset'}
\label{section:creating_comics_text_plus}

\begin{table}
    \caption{Some pairs of overlapping text boxes in the third comic series of the COMICS dataset \cite{iyyer2017amazing}. The one with a smaller area within the pairs is filtered, and the other is kept. As expected, most of their characters are shared.}
    \centering
    \resizebox{\textwidth}{!}{
        \begin{tabular}{@{}llllllll@{}}
            \toprule
            Page No & Panel No & Textbox No & Text                                  & x1 & y1  & x2  & y2  \\ \midrule
            5       & 2        & 0          & i quickly drag the yorn body into ... & 21 & 17  & 326 & 165 \\
            5       & 2        & 1          & i quickly drag the yorn body into ... & 24 & 21  & 326 & 165 \\
            20      & 0        & 1          & lovers in all the dark corners ...    & 1  & 0   & 410 & 132 \\
            20      & 0        & 0          & lovers in all the dark corners ...    & 0  & 0   & 410 & 121 \\
            30      & 2        & 2          & not know the fish had been fed        & 0  & 529 & 384 & 587 \\
            30      & 2        & 3          & not know ! the fish had been fed      & 0  & 540 & 384 & 587 \\ \bottomrule
        \end{tabular}}
    \label{table:overlapping_textboxes}
\end{table}

\begin{figure}[htbp!]
    \centering
    \includegraphics[width=0.1\textwidth]{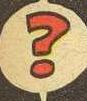}
    \includegraphics[width=0.1\textwidth]{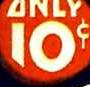}
    \includegraphics[width=0.25\textwidth]{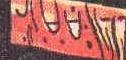}
    \includegraphics[width=0.1\textwidth]{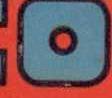}
    \caption{Some examples of textboxes are filtered from our COMICS Text+ dataset.}
    \label{fig:filtered_comics_ocr}
\end{figure}

After determining and training the best models for OCR, we applied them to the images cropped from panels using the coordinates of the textboxes provided in the COMICS dataset \cite{iyyer2017amazing}. However, the number of final textboxes with texts in our COMICS Text+ is different from the COMICS dataset for several reasons. Firstly, we filtered out advertisement panels and their textboxes, as well as textboxes that did not have matched texts, from our dataset. Secondly, some of our inferences resulted in erroneous or empty texts, which were also removed in order to improve the overall quality of the dataset. The statistics of the textboxes filtered for these reasons can be seen in Table \ref{table:filtered_comics_ocr_stats}, and some examples can be seen in Figure \ref{fig:filtered_comics_ocr}. After this step, we were left with 2,208,006 textboxes.

As a final preprocessing step, we filtered out textboxes that overlapped. In the COMICS dataset, some textboxes had highly intersecting textboxes, which can lower the quality of the dataset. We removed the smaller textbox in cases where they had more than 0.8 Intersection over Union (IoU) within the same panel, as shown in Table \ref{table:overlapping_textboxes}. This step reduced the total number of textboxes to 2,188,853, with 1,702,140 dialogue and 486,713 narratives. In comparison, the COMICS dataset has 2,545,728 textboxes. We believe that this preprocessing and filtering improves the overall quality of the dataset.

\begin{table}
    \caption{Statistics describing the filtered textboxes that are part of the COMICS dataset \cite{iyyer2017amazing} but not in COMICS Text+, by the result of empty or erroneous inferences. The majority of the images are images that are not textboxes, and the others that increase character and word average, are misdetections of textboxes in advertisement pages.}
    \centering
    \begin{tabular}{@{}ccccc@{}}
        \toprule
        textbox count & char. avg. & word avg. & char. median & word median \\ \midrule
        9508          & 6.43       & 2.12      & 3            & 1           \\ \bottomrule
    \end{tabular}
    \label{table:filtered_comics_ocr_stats}
\end{table}

\subsection{Postprocessing on Raw OCR Output}

We applied two postprocessing approaches to the extracted texts. The first was a rule-based approach, and the second was a neural spell-checking approach. However, the second approach did not improve the overall quality of the texts (see subsection \ref{subsection:neural_spell-checking_attempts}).

We improved the process described in \cite{iyyer2017discourse} for the rule-based approach by adding several additional steps. In Iyyer's work, two types of mistakes were targeted: mistakes in recognizing the first letter of a word (e.g., "eleportation" instead of "teleportation"), and errors in single characters. While these types of errors can be introduced by the OCR system, we believe that some are also the result of faulty textbox detection coordinates. As a result, we extended and modified their post-processing approach by considering errors that include not just the first letter but also letters from starting or ending.

\begin{algorithm}[H]
    \caption{Our algorithm for creating a dictionary for replacing tokens in our OCR output with PyEnchant suggestions. }
    \label{alg:pyenchant_replacement_dict}
    \begin{algorithmic}
        \Procedure{CreateTokenReplacementDict}{}
        \State $corpus \gets set(nltk\_corpus + brown\_corpus)$
        \ForEach {$token \in tokens$}

        \If{$token.length \geq 4  \AND token \notin corpus \AND  \Call{CheckPyEnchant}{$token$}$}
        \State $suggestions \gets \Call{GetSuggestions}{$token$}$

        \ForEach {$suggestion \in suggestions$}
        \If{$\Call{SuggestionStartsWith}{$token$} \OR  \Call{SuggestionEndsWith}{$token$}$}
        \State $\Call{AddTokenReplacementDict}{$token, suggestion$}$
        \EndIf
        \EndFor

        \EndIf
        \EndFor
        \EndProcedure
    \end{algorithmic}
\end{algorithm}

\comm{
    For each of these words that is length three or longer, we look up the most likely suggestion offered by PyEnchant. If the only difference between the most likely suggestion and the original word is an additional letter in the first position of the suggestion, then we replace the word with the suggestion everywhere in our corpus.
}

\comm{
    7) nltk corpus words + nltk.corpus.brown.words set yap => brown u niye seçtiğini anlat
    8) word frequency deki 10000 ile 100000 aralığındaki kelimeleri al
    9) kelime pyenchant check false ise, 3 chardan büyükse, corpuslarda yoksa
    10) pyenchant suggestionları al
    11) eğer suggestionlar bu kelime ile başlıyorsa ya da bitiyorsa
    12) replacement dict e ekle
    13) save it
}

To address the first type of error, we tokenized the OCR output and created a vocabulary of tokens sorted by the number of occurrences in descending order. We used the NLTK Punkt Tokenizer and Word Tokenizer\footnote{\url{https://www.nltk.org/}} for tokenization. We then applied Algorithm \ref{alg:pyenchant_replacement_dict} to tokens ranked between 10,001 and 100,000, assuming that misspelled words are less frequent. This allowed us to create a dictionary that could be used to replace those tokens in the output.

Our algorithm differs from Iyyer's \cite{iyyer2017discourse} approach in several ways. Firstly, we check whether the tokens are part of a corpus, using NLTK's default corpus and the Brown Corpus \cite{brown_corpus-francis1979brown} specifically. This is because the Golden Age of comics and the publication date of the Brown Corpus are closer in time than other corpora, which date from the 1950s and 1961. Secondly, instead of checking the token difference for the first letter of PyEnchant suggestions, we check whether the suggestion starts or ends with the token. This allows us to fix errors introduced by textbox detection errors.

\begin{algorithm}[H]
    \caption{Our algorithm for applying rule-based post-processing to OCR output to finalize our COMICS Text+ dataset.}
    \label{alg:apply_post_processing}
    \begin{algorithmic}
        \Procedure{ApplyPostProcessing}{}
        \State $valid\_characters \gets \{ a, d, i, m, s, t, x \}$
        \ForEach {$token \in all_tokens$}
        \IfNot{$\Call{CheckIfThereIsPuncBeforeOrAfter}{$token$}$}
        \State $token \gets \Call{ReplaceTokenIfInReplacementDict}{$token$}$
        \If{$token.length = 1  \AND token \notin valid\_characters$}
        \State \Call{Remove}{$token$}
        \EndIf
        \EndIf
        \EndFor
        \EndProcedure
    \end{algorithmic}
\end{algorithm}

\comm{
    1) NLTK’s Punkt Tokenizer
    2) tokenize sentences
    3) tokenize sentences into words
    4) create word frequency dictionary
    5) sort them by most occuring to least
    6) save the word frequency

    7) nltk corpus words + nltk.corpus.brown.words set yap => brown u niye seçtiğini anlat
    8) word frequency deki 10000 ile 100000 aralığındaki kelimeleri al
    9) kelime pyenchant check false ise, 3 chardan büyükse, corpuslarda yoksa
    10) pyenchant suggestionları al
    11) eğer suggestionlar bu kelime ile başlıyorsa ya da bitiyorsa
    12) replacement dict e ekle
    13) save it

    14) tüm datayı oku
    15) tokenize by words
    16) apply pyenchant postprocessing
    eğer kelimeden önce ya da noktalama işareti yoksa ve replacement dict te ise değiştir. => primarily for '-' symbol
    17) apply single char postprocessing
    kelime valid single char değilse öncesinde ya da sonrasında noktalama işareti yoksa wordu(characteri) sil. (f.b.i. gibi örnekleri korumak adına)

}

Erroneous single characters are detected by checking whether they have punctuation marks before or after them. If they do not have punctuation marks, we check whether they are in the set of valid single characters, which includes \textit{a, d, i, m, s, t, x}. These characters frequently occur in the text and are likely to be part of a word. We check for punctuation marks because there may be abbreviations of special names in the text (e.g., \textit{F.B.I.}, \textit{U.S.A.}). We then apply Algorithm \ref{alg:apply_post_processing} to post-process all of our OCR output and prepare it for its final form.

Our dataset shares both the raw OCR output and the post-processed version. A total of 110,838 textboxes were affected by the first part of our post-processing, where some of their tokens were replaced by PyEnchant. A total of 131,833 textboxes were updated with the second part of the process, where some of their single-character tokens were deleted. Both of these post-processing steps affected 18,580 textboxes as their intersections. It is difficult to measure the effects of the rule-based post-processing quantitatively, as only a few GT samples are affected.

\subsection{Neural Spell Checking Attempts}
\label{subsection:neural_spell-checking_attempts}

\begin{table}
    \caption{Evaluation of Contextual Spell Check contextual spell correction with respect to COMICS Dataset OCR and COMICS Text+ Data. 270 out of 500 ground truth data is used because that much of the data is considered and, as a result, corrected by the Contextual Spell Check.}
    \centering
    \resizebox{\textwidth}{!}{
        \begin{tabular}{@{}cccccc@{}}
            \toprule
                               & \textbf{Char. Recall} & \textbf{Char. Precision} & \textbf{Word Acc.} & \multicolumn{1}{p{3.2cm}}{\centering \textbf{Word Acc. Ignore Symbol}} & \textbf{1-N.E.D} \\ \midrule
            COMICS Dataset OCR & 0.9235                & 0.9654                   & 0.1037             & 0.4741                                                                 & 0.9312           \\
            COMICS Text+       & 0.9698                & 0.9739                   & 0.3407             & 0.6519                                                                 & 0.9779           \\
            Contextual Spell Check                                                                                                                                                                 \\ \cite{Goel_Contextual_Spell_Check_2021} on COMICS Text+ & 0.8523                & 0.9185                   & 0.0111             & 0.0222                           & 0.8078           \\ \bottomrule
        \end{tabular}}
    \label{table:contextual_spell_checking_results}
\end{table}

In addition to the rule-based post-processing, we also applied neural spell-checking and correction using NeuSpell \cite{jayanthi-etal-2020-neuspell}, and Contextual Spell Check \cite{Goel_Contextual_Spell_Check_2021}. Contextual Spell Check first identifies out-of-vocabulary words and uses BERT \cite{devlin2019bert} to generate predictions for those words. However, this approach significantly decreases the accuracy of our OCR output (see Table \ref{table:contextual_spell_checking_results}). We suspect that the dictionary used by the tool is not comprehensive or adequate for our domain of Golden Age comics, which leads to a decline in performance.

\begin{table}
    \caption{Evaluation of Neuspell \cite{jayanthi-etal-2020-neuspell} neural spelling correction toolkit's \currier{BERT Checker} with respect to COMICS Dataset OCR and COMICS Text+ Data. 273 out of 500 ground truth data is used because that much of the data is considered and, as a result, corrected by the Neuspell toolkit.}
    \centering
    \resizebox{\textwidth}{!}{
        \begin{tabular}{@{}cccccc@{}}
            \toprule
                               & \textbf{Char. Recall} & \textbf{Char. Precision} & \textbf{Word Acc.} & \multicolumn{1}{p{3.2cm}}{\centering \textbf{Word Acc. Ignore Symbol}} & \textbf{1-N.E.D} \\ \midrule
            COMICS Dataset OCR & 0.9207                & 0.9664                   & 0.1062             & 0.4799                                                                 & 0.9286           \\
            COMICS Text+       & 0.9719                & 0.9733                   & 0.2821             & 0.6227                                                                 & 0.9753           \\
            Neuspell \cite{jayanthi-etal-2020-neuspell}                                                                                                                                            \\ on COMICS Text+ & 0.9525                & 0.95                     & 0.0147             & 0.2564                                                                 & 0.9354           \\ \bottomrule
        \end{tabular}}
    \label{table:neuspell_eval}
\end{table}

NeuSpell uses neural models for spelling correction, which are trained by posing the task as a sequence labeling task. In this task, a correct word is tagged with itself, and a misspelled token is marked with its suitable correction. We used the "BERT Checker" model from their trained models. Although it performed better than Contextual Spell Check, it still reduced the quality of our OCR output (see Table \ref{table:neuspell_eval}). Both of these tools marked different texts in the textboxes for potential changes. To compare their performance on the same data, we selected the texts corrected by both tools and compared their results to the COMICS dataset OCR and our raw OCR output. Our raw OCR output outperformed all other methods (see Table \ref{table:neuspell_vs_contextual_eval}).

\begin{table}
    \caption{Evaluation of Contextual Spell Check \cite{Goel_Contextual_Spell_Check_2021} contextual spell correction and Neuspell \cite{jayanthi-etal-2020-neuspell} with respect to COMICS Dataset OCR and COMICS Text+ Data. 189 out of 500 ground truth data is used because that much of the data is considered and, as a result, corrected by both of the neural spell checkers.}
    \centering
    \resizebox{\textwidth}{!}{
        \begin{tabular}{@{}cccccc@{}}
            \toprule
                               & \textbf{Char. Recall} & \textbf{Char. Precision} & \textbf{Word Acc.} & \multicolumn{1}{p{3.2cm}}{\centering \textbf{Word Acc. Ignore Symbol}} & \textbf{1-N.E.D} \\ \midrule
            COMICS Dataset OCR & 0.9124                & 0.9607                   & 0.0952             & 0.4656                                                                 & 0.9271           \\
            COMICS Text+       & 0.9654                & 0.9679                   & 0.2646             & 0.5873                                                                 & 0.9769           \\
            Contextual Spell Checker                                                                                                                                                               \\ \cite{Goel_Contextual_Spell_Check_2021} on COMICS Text+ & 0.8465                & 0.9131                   & 0.0106             & 0.0159                                                                 & 0.8104           \\
            Neuspell \cite{jayanthi-etal-2020-neuspell}                                                                                                                                            \\ on COMICS Text+                 & 0.9426                & 0.9414                   & 0.0212             & 0.1534                                                                 & 0.9273           \\ \bottomrule
        \end{tabular}}
    \label{table:neuspell_vs_contextual_eval}
\end{table}

\section{Results \& Discussion}
\label{section:res_exp}

\subsection{Finalized Dataset Results}
\label{res:finalized_dataset_results}

\begin{table}
    \caption{Final comparison of our dataset called \currier{COMICS Text+} and COMICS dataset \cite{iyyer2017amazing} OCR results, measured on ground truth data. Our dataset outperforms the COMICS dataset OCR in all metrics. Our dataset increases word accuracy by 337\% and word accuracy by ignoring symbols by 140\% on the speech bubbles of the COMICS dataset.}
    \centering
    \resizebox{\textwidth}{!}{
        \begin{tabular}{@{}cccccc@{}}
            \toprule
                                  & \textbf{Char. Recall} & \textbf{Char. Precision} & \textbf{Word Acc.} & \multicolumn{1}{p{3.2cm}}{\centering \textbf{Word Acc. Ignore Symbol}} & \textbf{1-N.E.D} \\ \midrule
            COMICS Dataset OCR    & 0.9293                & 0.9718                   & 0.1301             & 0.516                                                                  & 0.9302           \\
            \textbf{COMICS Text+} & \textbf{0.9747}       & \textbf{0.9797}          & \textbf{0.4051}    & \textbf{0.7228}                                                        & \textbf{0.9774}  \\ \bottomrule
        \end{tabular}}
    \label{table:final_results_comics_vs_ours}
\end{table}

Table \ref{table:final_results_comics_vs_ours} compares the OCR results of COMICS \cite{iyyer2017amazing} and our method, COMICS Text+. Our method outperforms COMICS on all evaluation metrics, particularly character recall and word accuracy. This can be seen in the qualitative comparison in Table \ref{table:comics_text_plus_vs_comics}.

We attribute these results to using labeled data specific to the comic book domain and fine-tuning state-of-the-art text detection and recognition models. As shown in Figure \ref{fig:best_det_recog_data_size_gt}, the initial performance of these models is poor on the comic book data due to the unique characteristics of comic book text, such as speech bubbles and varied fonts. This demonstrates the need for fine-tuning using domain-specific data to achieve accurate OCR results in the comic book domain. Although labeling data is time-consuming, we show in Figures \ref{fig:best_det_data_size} and \ref{fig:best_recog_data_size} that even a limited amount of data (around 400 images) can improve the performance of SOTA models significantly. This study highlights the importance of both qualitative and quantitative analysis in evaluating OCR datasets and the need for domain-specific data and fine-tuning models to achieve accurate results in the comic book domain.

\begin{table}[htbp!]
    \caption{Qualitative comparison between samples of the COMICS Text+ dataset and COMICS dataset OCR. The examples shown are selected by their difference in 1-N.E.D. metric from ground truth data.}
    \centering
    \resizebox{\textwidth}{!}{
        \begin{footnotesize}
            \begin{tabular}{ | c | m{3cm} | m{3cm} | m{3cm} |  }
                \hline

                \begin{normalsize}
                    \textbf{Speech Bubble}
                \end{normalsize}
                 &
                \begin{normalsize}
                    \textbf{COMICS Dataset OCR}
                \end{normalsize}
                 &
                \begin{normalsize}
                    \textbf{COMICS Text+}
                \end{normalsize}
                 &
                \begin{normalsize}
                    \textbf{Ground Truth}
                \end{normalsize}

                \\ \hline

                \begin{minipage}{.3\textwidth}
                    \includegraphics[width=\linewidth]{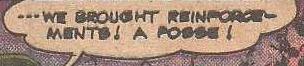}
                \end{minipage}
                 &
                we seou$\Phi$ht reinforce
                 &
                - - - we brought reinforce - ments ! a poose !
                 & - - - we brought reinforce - ments ! a posse !
                \\
                \hline

                \begin{minipage}{.3\textwidth}
                    \includegraphics[width=0.6\linewidth]{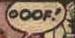}
                \end{minipage}
                 &
                ooor ! 000f ,
                 &
                ooof !
                 & ooof !
                \\
                \hline

                \begin{minipage}{.3\textwidth}
                    \includegraphics[width=0.5\linewidth]{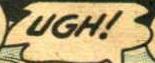}
                \end{minipage}
                 &
                l / gh / ugh
                 &
                ugh !
                 & ugh !
                \\
                \hline

                \comm{
                    \begin{minipage}{.3\textwidth}
                        \includegraphics[width=0.5\linewidth]{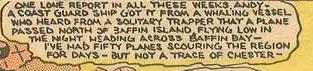}
                    \end{minipage}
                 &
                    who heard from a solitary trapper thayaplane passed north 3f bafpin island flyna low in ive mab fifty planes scouring the region for days but not a trace of chester
                 &
                    one lou - report - all these weeks andy - a coast guard ship got it from a whaling vessel who heard from > youthry trapper that a plane passed north - - buffin island ' flywa low in the night heading acrond baffin day - i ' ve nab fifty planes scouring the region for days but not a trace of chester -
                 & one lone report in all these weeks , andy - a coast guard ship got it from a whaling vessel who heard from a solitary trapper that a plane passed north of baffin island flying low in the night heading across baffin bay - i ' ve had fifty planes scouring the region for days - but not a trace of chester -
                    \\
                    \hline
                }

                \begin{minipage}{.3\textwidth}
                    \includegraphics[width=0.6\linewidth]{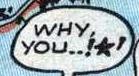}
                \end{minipage}
                 &
                why ,
                 &
                why you . . . i
                 & why you . . ! * '
                \\
                \hline

                \begin{minipage}{.3\textwidth}
                    \includegraphics[width=0.6\linewidth]{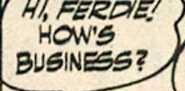}
                \end{minipage}
                 &
                hon ' s business ?
                 &
                hi , ferdie ! how ' s business ?
                 & hi , ferdie ! how ' s business ?
                \\
                \hline

                \begin{minipage}{.3\textwidth}
                    \includegraphics[width=\linewidth]{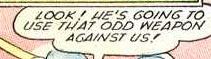}
                \end{minipage}
                 &
                c / se that odd weapon against us
                 &
                look , lie ' s going to use that odd weapon against us !
                 & look , he ' s going to use that odd weapon against us !
                \\
                \hline

                \begin{minipage}{.3\textwidth}
                    \includegraphics[width=0.7\linewidth]{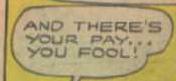}
                \end{minipage}
                 &
                and there ' s
                 &
                and there ' s your pay . . . you
                 & and there ' s your pay . . . you fool !
                \\
                \hline

                \begin{minipage}{.3\textwidth}
                    \includegraphics[width=0.5\linewidth]{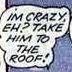}
                \end{minipage}
                 &
                am crazy ea take him
                 &
                im crazy , eh ? take him to the roof !
                 & im crazy , eh ? take him to the roof !
                \\
                \hline

                \begin{minipage}{.3\textwidth}
                    \includegraphics[width=0.8\linewidth]{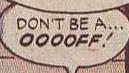}
                \end{minipage}
                 &
                don ' t be a ...
                 &
                don ' t be a . . ooooff
                 & don ' t be a . . ooooff !
                \\
                \hline

                \begin{minipage}{.3\textwidth}
                    \includegraphics[width=0.7\linewidth]{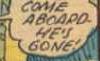}
                \end{minipage}
                 &
                aboard . gone
                 &
                come aboard - he ' s gone !
                 & come aboard - he ' s gone !
                \\
                \hline

                \begin{minipage}{.3\textwidth}
                    \includegraphics[width=0.5\linewidth]{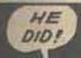}
                \end{minipage}
                 &
                did !
                 &
                he did !
                 & he did !
                \\
                \hline

                \begin{minipage}{.3\textwidth}
                    \includegraphics[width=\linewidth]{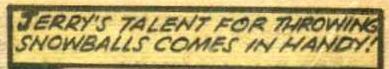}
                \end{minipage}
                 &
                erry ' s 74lew\(<\)unk\(>\)acr throwhag forma snowballs comer navamdy -
                 &
                terry ' s talent for throwing snowballs comes in handy !
                 & jerry ' s talent for throwing snowballs comes in handy !
                \\
                \hline
            \end{tabular}
        \end{footnotesize}}
    \label{table:comics_text_plus_vs_comics}
\end{table}

\subsubsection{Error Analysis}

Table \ref{table:comics_text_plus_ocr_flaws} presents the most common errors in the COMICS TEXT+ dataset. These errors include:

\begin{itemize}
    \item Difficulty detecting and recognizing symbols, particularly commas and dots.
    \item Text detection errors due to uncommon textures and background colors, as shown in the first and third examples of Table \ref{table:comics_text_plus_ocr_flaws}.
    \item Poorly detected bounding boxes lead to OCR issues. Speech bubble and narrative box detection or segmentation should be solved before the OCR step. Repercussions of this issue can lead to totally omitted or partially detected words, word groups, or multiple partial textboxes.
\end{itemize}

To address the issues with bounding boxes, improved textbox detection or segmentation methods may be necessary. However, addressing the other problems will likely require more extensive text detection and recognition datasets for fine-tuning.

\begin{table}[htbp!]
    \caption{Results of qualitative error analysis on COMICS Text+ dataset. The examples shown are selected from the least performing speech bubbles based on the 1 - N.E.D. metric from ground truth data. The COMICS Dataset OCR results for the selected speech bubbles are also added for reference.}
    \centering
    \resizebox{\textwidth}{!}{
        \begin{footnotesize}
            \begin{tabular}{ | c | m{3cm} | m{3cm} | m{3cm} |  }
                \hline

                \begin{normalsize}
                    \textbf{Speech Bubble}
                \end{normalsize}
                 &
                \begin{normalsize}
                    \textbf{COMICS Dataset OCR}
                \end{normalsize}
                 &
                \begin{normalsize}
                    \textbf{COMICS Text+}
                \end{normalsize}
                 &
                \begin{normalsize}
                    \textbf{Ground Truth}
                \end{normalsize}

                \\ \hline
                \begin{minipage}{.3\textwidth}
                    \includegraphics[width=\linewidth]{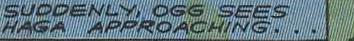}
                \end{minipage}
                 &
                suddenly , ogg sees haga approaching
                 &
                sudo ogg
                 & suddenly , ogg sees haga approaching . . .
                \\
                \hline

                \begin{minipage}{.3\textwidth}
                    \includegraphics[width=0.5\linewidth]{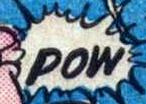}
                \end{minipage}
                 &
                pow
                 &
                ma dow
                 & pow
                \\
                \hline

                \begin{minipage}{.3\textwidth}
                    \includegraphics[width=\linewidth]{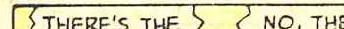}
                \end{minipage}
                 &
                there ' s the no , the
                 &
                no the there ' s the
                 & there ' s the no , the
                \\
                \hline

                \begin{minipage}{.3\textwidth}
                    \includegraphics[width=\linewidth]{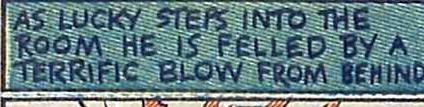}
                \end{minipage}
                 &
                asluckx ste / s - int the room he i spelledby a terrific blow from behind
                 &
                into felled terrific blow from
                 & as lucky steps into the room he is felled by a terrific blow from behind
                \\
                \hline

                \begin{minipage}{.3\textwidth}
                    \includegraphics[width=\linewidth]{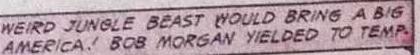}
                \end{minipage}
                 &
                weird jungle beast pould bring a big america bob morgan yelded 7 tema
                 &
                weird jungle beast would bring a big yielded to temp . america . ! bob morgan
                 & weird jungle beast would bring a big america ! bob morgan yielded to temp .
                \\
                \hline

                \begin{minipage}{.3\textwidth}
                    \includegraphics[width=\linewidth]{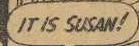}
                \end{minipage}
                 &
                itis susan
                 &
                it ! susan !
                 & it is susan !
                \\
                \hline

                \begin{minipage}{.3\textwidth}
                    \includegraphics[width=0.8\linewidth]{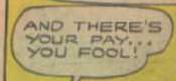}
                \end{minipage}
                 &
                and there ' s
                 &
                and there ' s your pay . . . you
                 & and there ' s your pay . . . you fool !
                \\
                \hline

                \begin{minipage}{.3\textwidth}
                    \includegraphics[width=\linewidth]{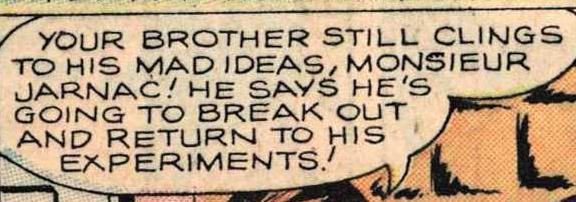}
                \end{minipage}
                 &
                your brother still clings to his mad ideas , monsieur uarnac ! he says he ' s going to break out and return to his experiments .
                 &
                your brother still clings monsieur to his mad ideas , jarnac ! he says he ' s going to break out and return to his experiments , !
                 & your brother still clings to his mad ideas , monsieur jarnac ! he says he ' s going to break out and return to his experiments !
                \\
                \hline

                \begin{minipage}{.3\textwidth}
                    \includegraphics[width=0.6\linewidth]{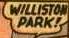}
                \end{minipage}
                 &
                williston parki
                 &
                willist . park !
                 & williston park !
                \\
                \hline

                \begin{minipage}{.3\textwidth}
                    \includegraphics[width=0.6\linewidth]{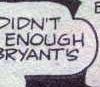}
                \end{minipage}
                 &
                didnt enough ryants
                 &
                i didn ' t enough suryant ' s
                 & didn ' t enough bryant ' s
                \\
                \hline
            \end{tabular}
        \end{footnotesize}}
    \label{table:comics_text_plus_ocr_flaws}
\end{table}

\subsection{Results with Cloze Style Tasks}

We have reproduced the models and experimental setup described in \cite{iyyer2017amazing} and present our results in Table \ref{table:comics_reproduction_results}. For most of the experiment cases, our reproduction results are similar to those reported by a low margin. Given our reproduction results and experiment setup of character coherence, the validity of character coherence can be argued. Although the task is sound, the implementation shows us it only evaluates whether a text is before or after another text since no explicit character encoding or feature is provided. We think the overall difference in results can stem from the dataset generation since the dataset itself is not shared. Although a script for generating the dataset is shared, it does not include explicit instructions on setting the hyperparameters. This may contribute to the discrepancies between our reproduction results and those reported in the original study.

\begin{figure}[htbp!]
    \centering
    \includegraphics[width=\textwidth]{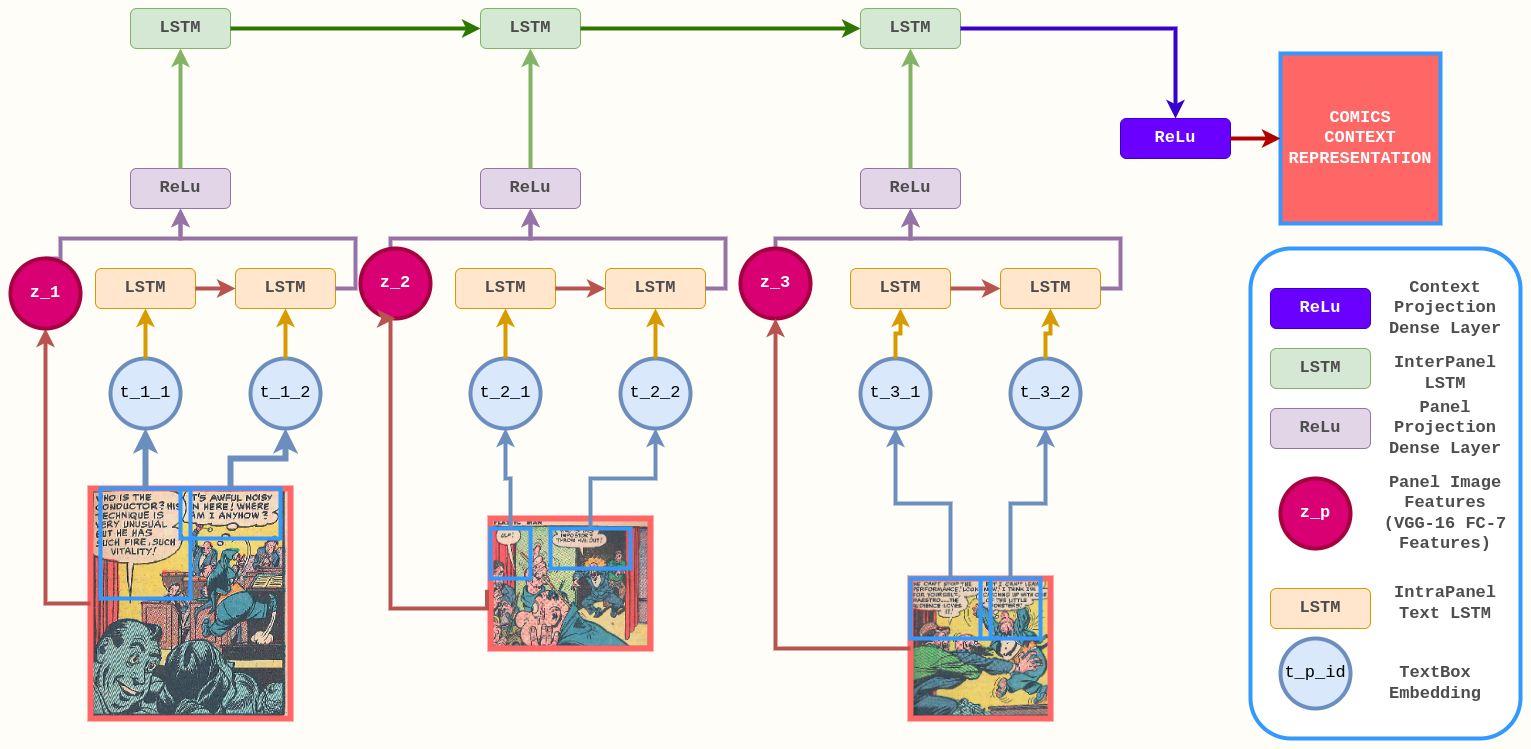}
    \caption{Context representation extraction from the model's architecture using both image and text modality. It is called \textbf{image-text} architecture. Single modality architectures are named as \textbf{image-only} and \textbf{text-only}. Those models keep the same structure, disregarding the processing of other modalities.}
    \label{fig:comics_context_model}
\end{figure}

Iyyer et al. \cite{iyyer2017amazing} propose three tasks (text cloze, visual cloze, and character coherence) to evaluate a model's ability to understand the closure of a comic book narrative based on information from the previous three panels. A model that performs well on these tasks is able to make inferences between panel transitions and estimate the features of the following panel. This can be seen as a measure of a model's ability to understand the overall narrative structure of a comic book.

\begin{table}[htbp!]
    \caption{Comparison of our replication results Iyyer et al.'s \cite{iyyer2017amazing} results based on accuracy. Human baseline is also shared for only hard tasks since the easy task is trivial for humans. \textit{NC-Image-text} models use only the final panel's image without any information from preceding context panels. No context model results strike the usefulness of the context for the cloze style tasks and narrative understanding.}
    \centering
    \resizebox{\textwidth}{!}{
        \begin{tabular}{@{}ccccccccccccc@{}}
            \toprule
            MODEL                            & \multicolumn{4}{c}{{\ul \textbf{Text Cloze}}} & \multicolumn{4}{c}{{\ul \textbf{Visual Cloze}}} & \multicolumn{2}{c}{{\ul \textbf{Char. Coherence}}}                                                                                                                 \\
            \multirow{2}{*}{Task Difficulty} & \multicolumn{2}{c}{EASY}                      & \multicolumn{2}{c}{HARD}                        & \multicolumn{2}{c}{EASY}                           & \multicolumn{2}{c}{HARD} & \multicolumn{2}{c}{}                                                               \\
                                             & Original                                      & Ours                                            & Original                                           & Ours                     & Original               & Ours                  & Original & Ours & Original & Ours \\ \midrule
            Text-only                        & 63.4                                          & 62.8                                            & 52.9                                               & 51.7                     & 55.9                   & 51.8                  & 48.4     & 45.0 & 68.2     & 70.9 \\
            Image-only                       & 51.7                                          & 50.7                                            & 49.4                                               & 45.6                     & 85.7                   & 79.5                  & 63.2     & 56.1 & 70.9     & 70.7 \\
            Image-text                       & 68.6                                          & 62.1                                            & 61.0                                               & 57.2                     & 81.3                   & 83.6                  & 59.1     & 59.5 & 69.3     & 71.5 \\
            NC-Image-text                    & 63.1                                          & 57.3                                            & 59.6                                               & 56.5                     & \multicolumn{2}{c}{-}  & \multicolumn{2}{c}{-} & 65.2     & 75.6                   \\ \midrule
            Human                            & \multicolumn{2}{c}{-}                         & \multicolumn{2}{c}{84}                          & \multicolumn{2}{c}{-}                              & \multicolumn{2}{c}{88}   & \multicolumn{2}{c}{87}                                                             \\ \bottomrule
        \end{tabular}}
    \label{table:comics_reproduction_results}
\end{table}

The general setup for the tasks involves using the features of context panels $p_{i-3}, p_{i-2}, p_{i-1}$ as input for a model that predicts the features of panel $p_i$ from a single modality. This setup can be enhanced by using an arbitrary number of context panels, but that is a topic for future study.

The final panel's features are predicted using a cloze-style framework. For all experiments, three candidates are presented, and the models use the extracted context information to score the correct candidate higher than the others. Figure \ref{fig:comics_context_model} illustrates the context extraction process. The three tasks are as follows:

\begin{itemize}
    \item The \textbf{Text Cloze} task involves predicting the text of the final panel $p_i$. To reduce the task's complexity and measure the models' ability to learn from inter-panel transitions, the final panels are selected from those with only a single text box. Architectures that make use of the visual modality, such as \textit{image-text} and \textit{image-only}, apply masking to the text box areas so that the model cannot "see" the text content and provides the artwork features along with the context to score the text box candidates.
    \item The \textbf{Visual Cloze} task involves predicting the artwork of the final panel $p_i$. Unlike the \textit{Text Cloze} task, the models are not provided with the text features of the final panel.
    \item The \textbf{Character Coherence} task involves reordering two dialogue text boxes for the final panel.
\end{itemize}

Each task has two variations based on difficulty. In the \textit{easy} case, the wrong candidates are selected from comic books that are different from the correct candidates. In contrast, the candidates of the \textit{hard} case come from the same comic book.


Using our implementation, the experiments' text data is replaced with \textit{COMICS TEXT+}. The test results of this training are shared in Table \ref{table:ctp_model_results}. Compared to our reproduction results using COMICS OCR data, an increase in accuracy can be observed for all text cloze task cases. We achieved state-of-the-art (SOTA) results using the same models compared to Iyyer et al.'s \cite{iyyer2017amazing}. However, there was no or only a slight improvement in the visual cloze tasks. This indicates that when working with this architecture, the text modality has limited effectiveness on visual cloze.

While the improvement in the text-cloze task is notable, it is not groundbreaking, even if it is SOTA. This suggests that the current model architecture and experimental setup have reached their maximum potential. To make further progress in the multimodal processing of comics, it is necessary to develop new model architectures and experimental setups better suited to the domain. Hopefully, we will see such approaches in the near future.

\begin{table}[htbp!]
    \caption{Our results with replacing COMICS OCR dataset with \textit{COMICS TEXT+} dataset using the same model architecture and same training procedure. Star indicates SOTA, single up arrow indicates a slight increase, and double up arrows indicate considerable improvement with respect to our replication results.}
    \centering
    \resizebox{\textwidth}{!}{
        \begin{tabular}{cccccc}
            \hline
            MODEL           & \multicolumn{2}{c}{{\ul \textbf{Text Cloze}}} & \multicolumn{2}{c}{{\ul \textbf{Visual Cloze}}} & {\ul \textbf{Char. Coherence}}                          \\
            Task Difficulty & EASY                                          & HARD                                            & EASY                           & HARD            &      \\ \midrule
            Text-only       & 64.5 $\upuparrows$ $\star$                    & 55.0  $\upuparrows$ $\star$                     & 51.6 $\uparrow$                & 45.5 $\uparrow$ & 67.0 \\
            Image-only      & 52.6  $\upuparrows$$\star$                    & 50.3  $\upuparrows$ $\star$                     & -                              & -               & -    \\
            Image-text      & 64.1  $\upuparrows$                           & 61.4  $\upuparrows$ $\star$                     & 83.5 $\star$                   & 59.0 $\star$    & 68.0 \\ \bottomrule
        \end{tabular}}
    \label{table:ctp_model_results}
\end{table}

\subsection{Performance on Modern Comics}

The models trained on COMICS Text+ are used to extract text from speech bubbles in "Modern Age" comics, specifically comics that were published after 2000. To do this, the inference procedure involves cropping the speech bubbles using our speech bubble detector and then running the models on the speech bubbles rather than the entire pages or panels. The same procedure would be applied to our dataset.

However, the performance turns out to be poor. Unfortunately, we cannot share any of the modern speech bubbles in our dataset due to copyright reasons. This drop in performance is primarily due to errors in the text detector, which is in line with our study findings. We found that text detectors are the bottleneck for OCR in comics. The model, as trained on our dataset, cannot separate text groups or words as well as in Golden Age comics.

Our publicly shared models can be used with Golden Age comics. However, the model will need to be retrained if they are to be used in comics from other periods or styles. In this study, we also found that a few hundred annotated data are sufficient for the models to perform close to their peak performance.

\section{Conclusion and Future Work}

We present the COMICS TEXT+ dataset, an improved version of the OCR data from the COMICS dataset \cite{iyyer2017amazing}. In contrast to the original COMICS dataset, we did not use any paid or proprietary software for OCR. Instead, we developed our own pipeline for generating text detection and text recognition data, which we used to fine-tune state-of-the-art deep learning-based text detection and recognition models to create an end-to-end OCR pipeline. We also split the dataset into two parts: \textit{COMICS Text+: Detection} and \textit{COMICS Text+: Recognition}. Finally, we applied rule-based post-processing to the OCR data to further augment the data. The resulting dataset contains more than 2 million transcriptions of textboxes and has 1 - N.E.D. of 0.9774, making it the most extensive and accurate dataset of its kind on this scale.

Our study shows that current computational approaches to understanding the narratives in comics are not yet able to achieve human-level performance, even when provided with high-quality text data. To fully exploit the multimodal nature of comics and create more generalizable models, it is important to access and process a diverse range of comic series from different periods, as this will provide more data for model training.

One particular challenge in this field is the lack of specialized text detection models for comics, which can be a bottleneck for OCR performance. Current text detection models may not provide satisfactory results when applied to different types of comics, so there is a need for the development of comics-specific text detection models.

In addition, we have demonstrated that current neural spell-checkers may not be effective in improving OCR quality in the context of comics. This highlights the need for the development of unified OCR models that are specifically tailored for this domain.

In this study, we focused on the extraction of dialogues and narratives from comics, but other studies have focused on other text modalities such as onomatopoeias (e.g., \cite{onomatopoeia_baek2022coo}). By extracting all text modalities from comics, we can gain a deeper understanding of the visual language used in these works.

Furthermore, it is important to accurately process low-level features of comics, such as text boxes, panels, characters, and other components. This will enable the development of more comprehensive comics processing backbones that can tackle high-level features of comics.

Overall, there is significant potential for further research in the field of computational studies of comics. By addressing the challenges of text detection, comics component processing, and unified OCR, we can enable the development of more effective computational methods for understanding the content of comics and enable the development of high-level processing methods for tasks such as narrative understanding and story generation.

\clearpage
\bibliographystyle{splncs04}
\bibliography{references}

\end{document}